\documentclass[pdflatex,sn-mathphys-num]{sn-jnl}

\usepackage{graphicx}%
\usepackage{multirow}%
\usepackage{amsmath,amssymb,amsfonts}%
\usepackage{amsthm}%
\usepackage{mathrsfs}%
\usepackage[title]{appendix}%
\usepackage{xcolor}%
\usepackage{textcomp}%
\usepackage{manyfoot}%
\usepackage{booktabs}%
\usepackage{algorithm}%
\usepackage{algorithmicx}%
\usepackage{algpseudocode}%
\usepackage{listings}%
\usepackage{float}

\usepackage{subcaption}
\usepackage{array}
\usepackage{comment}
\usepackage[most]{tcolorbox}
\usepackage{pifont} 
\usepackage{enumitem} 

\theoremstyle{thmstyleone}%
\usepackage{xcolor}

\theoremstyle{thmstyletwo}%

\theoremstyle{thmstylethree}%
\newtheorem{definition}{Definition}%
\newtheorem{hyp}{Hypothesis}%

\newtheorem{innertakeaway}{Takeaway}
\newenvironment{takeaway}
  {\begin{tcolorbox}[colback=gray!5, colframe=blue!40]\begin{innertakeaway}\ding{43} }
  {\end{innertakeaway}\end{tcolorbox}}
\begin{document}

\title[Article Title]{Can ``AI''  be a Doctor? A Study of Empathy, Readability, and Alignment in Clinical LLMs}


\author*[1]{\fnm{Mariano} \sur{Barone}}\email{mariano.barone@unina.it}

\author[1]{\fnm{Francesco} \sur{Di Serio}}\email{francesco.diserio@unina.it}
\equalcont{These authors contributed equally to this work.}

\author[2]{\fnm{Roberto} \sur{Moio}}\email{roberto.moio@unicampania.it}
\equalcont{These authors contributed equally to this work.}

\author[3]{\fnm{Marco} \sur{Postiglione}}\email{marco.postiglione@northwestern.edu}
\equalcont{These authors contributed equally to this work.}

\author[1]{\fnm{Giuseppe} \sur{Riccio}}\email{giuseppe.riccio3@unina.it}
\equalcont{These authors contributed equally to this work.}

\author[1]{\fnm{Antonio} \sur{Romano}}\email{antonio.romano5@unina.it}
\equalcont{These authors contributed equally to this work.}

\author[1]{\fnm{Vincenzo} \sur{Moscato}}\email{vmoscato@unina.it}
\equalcont{These authors contributed equally to this work.}

\affil*[1]{\orgdiv{Department of Electrical Engineering and Information Technology}, \orgname{University of Naples Federico II}, \orgaddress{\street{Via Claudio 21}, \city{Naples}, \postcode{80125}, \country{Italy}}}

\affil[2]{\orgdiv{Department of Translational Medical Sciences}, \orgname{University of Campania "Luigi Vanvitelli"}, \orgaddress{\street{Via Leonardo Bianchi}, \city{Naples}, \postcode{80131}, \country{Italy}}}

\affil[3]{\orgdiv{Department of Computer Science, McCormick School of Engineering and Applied Science}, \orgname{Northwestern University}, \orgaddress{\street{2309 Sheridan Rd}, \city{Evanston}, \postcode{60201}, \state{IL}, \country{United States}}}



\abstract{
Large Language Models (LLMs) are increasingly deployed in healthcare, yet their communicative alignment with clinical standards remains insufficiently quantified. We conduct a multidimensional evaluation of general-purpose and domain-specialized LLMs across structured medical explanations and real-world physician–patient interactions, analyzing semantic fidelity, readability, and affective resonance.
Baseline models amplify affective polarity relative to physicians (Very Negative: 43.14–45.10\% vs. 37.25\%) and, in larger architectures such as GPT-5 and Claude, produce substantially higher linguistic complexity (FKGL up to 16.91–17.60 vs. 11.47–12.50 in physician-authored responses). Empathy-oriented prompting reduces extreme negativity and lowers grade-level complexity (up to -6.87 FKGL points for GPT-5) but does not significantly increase semantic fidelity.
Collaborative rewriting yields the strongest overall alignment. Rephrase configurations achieve the highest semantic similarity to physician answers (up to $\mu = 0.93$) while consistently improving readability and reducing affective extremity. Dual stakeholder evaluation shows that no model surpasses physicians on epistemic criteria, whereas patients consistently prefer rewritten variants for clarity and emotional tone.
These findings suggest that LLMs function most effectively as collaborative communication enhancers rather than replacements for clinical expertise.
}

\keywords{Large Language Models, Healthcare AI, Empathy in AI, Readability, Medical Communication, MedQuAD, Medical Question-Answering}



\maketitle

\section{Introduction}\label{sec1}

As AI systems become increasingly capable and autonomous, ensuring their alignment with human values has become a practical concern rather than a purely theoretical one \cite{klingefjord2024humanvaluesalignai, röttger2025safetypromptssystematicreviewopen}. In patient-facing medical applications, misalignment may result in unclear communication, inappropriate reassurance, or unsafe recommendations, with direct consequences for patient trust and clinical decision-making.
This issue is particularly critical in healthcare, where Large Language Models (LLMs) are increasingly deployed in patient-facing settings, often in contexts characterized by vulnerability and emotional distress\cite{DBLP:journals/npjdm/ArmoundasL25}. Recent evidence shows that over 150{,}000 clinicians across more than 150 institutions already rely on AI-powered systems to assist with patient messaging\footnote{\url{https://www.nytimes.com/2024/09/24/health/ai-patient-messages-mychart.html}}, thus reshaping everyday clinical communication practices \cite{han2024ascleai, shool2025systematic, DBLP:journals/npjdm/RazaVK24}.

While existing research has largely focused on the factual accuracy of medical AI systems, particularly addressing hallucinations and reliability \cite{garciafernandez2025trustworthyaimedicinecontinuous,DBLP:journals/npjdm/AsgariBDKBYP25}, other determinants of effective clinical communication remain comparatively underexplored. In this work, we focus on three communicative dimensions that are central in clinical interactions: semantic correctness (preservation of medical meaning), readability (linguistic accessibility to non-expert users), and affective appropriateness (alignment of emotional tone with patient needs).
In the absence of these values, patients may experience confusion, anxiety, and erosion of trust in their healthcare providers \cite{wynia2010health, horvat2024barriers, ong1995doctor}. Despite the existence of well-established communication frameworks such as SPIKES \cite{baile2000spikes} and the Calgary-Cambridge Guide \cite{calgary2003marrying}, little empirical work has investigated whether modern LLMs reproduce or deviate from these communicative principles when interacting with patients\cite{DBLP:journals/npjdm/AgrawalCGJ25}.
We address this gap through a systematic evaluation of three communicative dimensions that are particularly relevant in clinical contexts: (1) \textit{semantic fidelity}, how faithfully AI responses match expert clinical judgment; (2) \textit{readability}, whether AI answers remain comprehensible across diverse literacy levels and cultural backgrounds; and (3) \textit{affective resonance}, the extent to which AI responses acknowledge patients' emotional needs. To this end, we conduct our analysis on a large-scale medical question-answering corpus comprising 47{,}457 entries derived from authoritative healthcare sources \cite{BenAbacha-TREC2017}, and investigate the following research questions:

\begin{enumerate}
    \item \textbf{RQ1 (\textit{Empathy}):} 
    How do LLMs compare to human physicians in expressing empathy and emotional awareness in clinical communication?
    \item \textbf{RQ2 (\textit{Readability}):} 
    Do LLM-generated responses differ from physician-authored answers in terms of linguistic readability?
    \item \textbf{RQ3 (\textit{Prompt-based Alignment}):} 
    Does the use of empathy-oriented prompting improve the emotional tone and readability of LLM outputs while preserving semantic fidelity?
    \item \textbf{RQ4 (\textit{Human-AI Collaboration}):} 
    Can LLMs improve the clarity and emotional appropriateness of physician-authored responses through collaborative rewriting?
    \item \textbf{RQ5 (\textit{Expert-Patient Value Alignment}):} 
    To what extent do different LLM configurations satisfy the distinct preferences expressed by medical experts and patients?
\end{enumerate}

Our study addresses these questions through one of the first large-scale empirical comparisons between LLM-generated and physician-authored medical communication. We identify systematic differences between models and human clinicians in emotional tone, readability, and semantic consistency, with no single approach consistently outperforming the others across all dimensions.

A central contribution of this work lies in the nature of the data examined. In contrast to prior studies that predominantly rely on social media content, patient self-reports, or synthetic dialogues, we analyze expert-authored medical communication produced by practicing clinicians in institutional settings. This enables an evaluation of alignment against high-standard clinical language as used in real healthcare workflows, rather than informal user-generated text. To the best of our knowledge, such data have been rarely leveraged in large-scale evaluations of LLMs for healthcare communication.

\section{Related Work}\label{sec2}

Empathy is widely recognized as a central component of effective clinical communication. In medical contexts, it is not merely a matter of emotional warmth but of calibrated emotional engagement, often described as \textit{detached concern} \cite{nembhard2023systematic}. Recent studies suggest that LLMs can generate emotionally resonant medical responses, although the empirical findings remain mixed. Ayers et al. \cite{ayers2023chatgpt} reported that 79\% of Reddit AskDocs users preferred ChatGPT responses over those written by physicians, largely due to perceived empathy and tone. However, the informal context of online forums limits the generalizability of these findings to real clinical practice.
More recently, analogous findings have been reported in oncological settings, where patients consistently rated chatbot responses as more empathetic than physician responses, further highlighting the divergence between lay and clinical perceptions of affective tone\cite{DBLP:journals/npjdm/ChenCPLLMEHCHFWR25}.
Similarly, Luo et al. \cite{luo2024assessing} proposed EMRank as a metric for quantifying empathy in LLM responses, reporting higher empathy scores for ChatGPT compared to physicians, though their analysis focused primarily on emotional expression rather than overall clinical adequacy.

Beyond emotional tone, readability and linguistic simplification represent further challenges in patient–provider communication. Roy et al. \cite{ROY2025108986} showed that GPT-5 can improve comprehension of medical information; similar benefits have been demonstrated when AI is used to simplify surgical consent forms through human-AI collaborative approaches \cite{DBLP:journals/npjdm/AliCTMJAGLSGGTSAZD24}, yet excessive simplification may reduce interpretability in complex clinical scenarios.

A primary mechanism through which these dimensions are implicitly adjusted is prompt engineering. Prompt engineering has emerged as a key modulator of LLM behavior in medical settings. Prior work shows that techniques such as chain-of-thought prompting improve reasoning transparency and task performance \cite{wei2022chain}, while role conditioning can increase perceived empathy in generated responses. However, few studies have operationalized established clinical communication frameworks such as SPIKES \cite{baile2000spikes} or the Calgary--Cambridge Guide \cite{calgary2003marrying} within prompt templates, limiting the transferability of these findings to structured clinical environments.

These variations in prompt design not only affect how models generate responses, but also complicate direct comparisons with physician-authored communication.
Several meta-analyses report that LLMs offer broader informational coverage, particularly in identifying symptoms, potential diagnoses, and treatment options, though concerns remain regarding precision and contextual appropriateness \cite{WangLi}. Model performance varies substantially across tasks, prompting strategies, and domain settings \cite{abrar2024empiricalevaluationlargelanguage}. In addition, commonly used benchmarks often rely on public forums or synthetic datasets, which lack domain realism and rarely include parallel expert-authored content.
These limitations have motivated a shift toward evaluation frameworks that better reflect clinical stakeholders and real-world deployment conditions.
\cite{ding2025aligninglargelanguagemodels,subramanian2024enhancing,DBLP:journals/npjdm/TamSKSPMOWVFMCSPW24}. Nevertheless, existing implementations remain limited in scale and scope, with relatively few studies integrating emotional, linguistic, and semantic dimensions within a unified evaluation framework\cite{DBLP:journals/npjdm/WangTYGGMWSLMJHMSJTWGYL26}.

Overall, the literature reveals persistent methodological limitations in assessing LLMs for clinical communication. Most evaluations adopt isolated or unidimensional perspectives, examining empathy, readability, or correctness separately rather than in combination. Many studies rely on small-scale human evaluations without rigorous inter-rater validation, which reduces statistical robustness and reproducibility. Moreover, limited control over semantic preservation complicates interpretation of whether improvements reflect genuine clinical adequacy or superficial stylistic variation. Our work seeks to address these gaps through a multidimensional evaluation framework that jointly examines emotional tone, readability, and semantic fidelity across multiple LLM configurations. In addition, we assess both AI-generated responses and LLM-assisted rewriting of physician-authored content, combining simulated expert assessment with human patient evaluation to capture distinct stakeholder perspectives.

\section{Methodology}\label{sec3}

\begin{figure*}[t]
    \centering
\includegraphics[width=\textwidth]{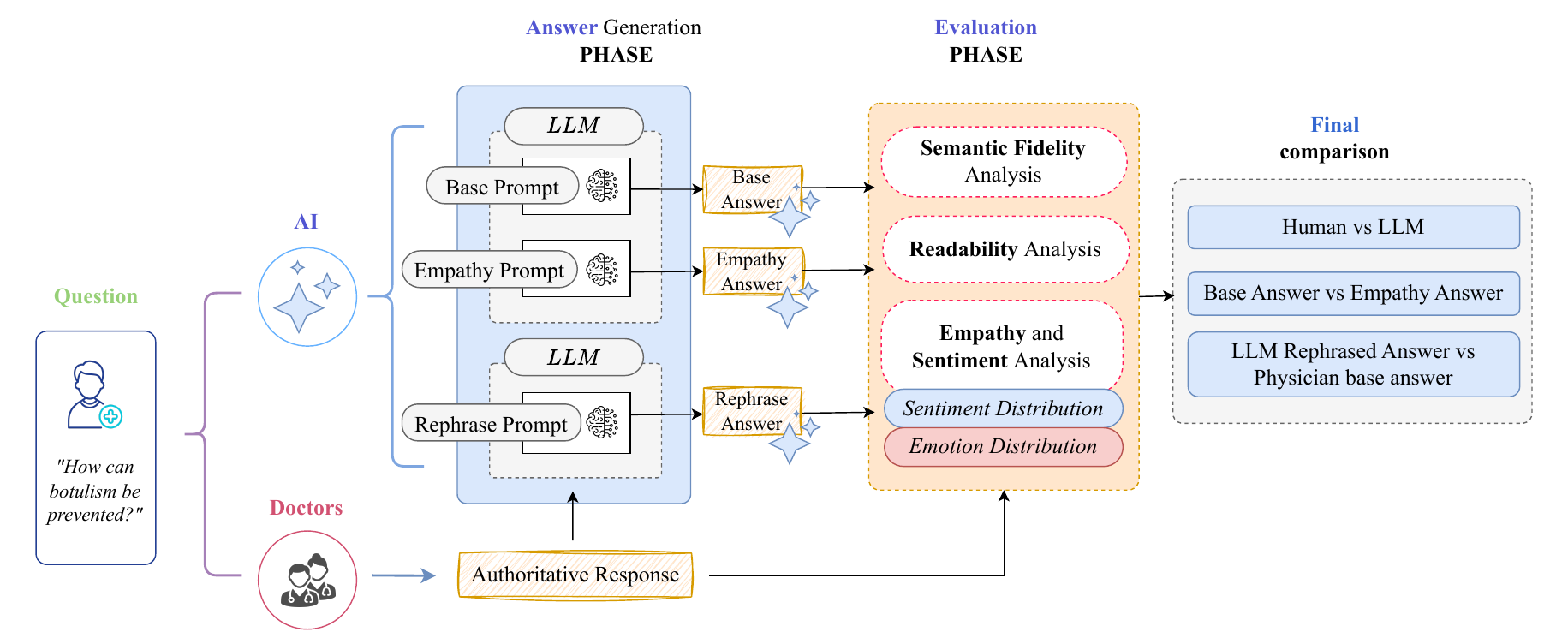}
    \caption{The framework compares LLM-generated and physician-authored answers across semantic similarity, readability, sentiment, and emotion. It includes both direct generation and LLM-based revision of expert responses, enabling evaluation of AI models as autonomous communicators and collaborative assistants in clinical settings.}
    \label{fig:workflow}
\end{figure*}

This section introduces a multidimensional evaluation framework designed to assess the case study across three core communicative dimensions: \textit{semantic fidelity}, \textit{readability}, and \textit{affective resonance (sentiment and empathy)}. The framework supports both the autonomous generation of responses by LLMs and the collaborative revision of expert-authored responses, simulating hybrid human–AI interaction scenarios. Figure~\ref{fig:workflow} illustrates the overall pipeline.

\subsection{Background}
Each communicative dimension is conceptually defined and formally operationalized using established computational metrics, as detailed below.

\begin{definition}[Semantic Fidelity]
    Let $r_h$ be the human-authored response and $r_m$ the model-generated response.  
    Let $\phi: \mathcal{T} \rightarrow \mathbb{R}^d$ be a sentence embedding function mapping a text sequence into a $d$-dimensional semantic space.  
    We define \textit{semantic fidelity} as
    \[
        \text{SF}(r_h, r_m) = \cos\big(\phi(r_h), \phi(r_m)\big),
    \]
    where cosine similarity quantifies the conceptual proximity between the embeddings.  
    Higher values indicate stronger alignment in global meaning.
\end{definition}

Semantic fidelity evaluates the degree to which model-generated responses preserve the conceptual content expressed by clinicians.  
We operationalize this dimension using cosine similarity computed over embeddings produced by the \texttt{BioBERT-mnli-snli-scinli-scitail-mednli-stsb} encoder.  
Because this metric does not capture fine-grained factual inaccuracies (e.g., incorrect dosages or omitted clinical entities), we interpret it as a measure of \emph{conceptual fidelity}, supplemented by domain-specific analyses reported in Section~\ref{sec5}.

\begin{definition}[Readability]
    Let $r$ be a text consisting of $W$ words, $S$ sentences, $Sy$ syllables, and $C$ complex words.  
    We define \textit{readability} as the vector
    \[
        \text{Read}(r) = 
        \big(
            \text{FKGL}(r),\;
            \text{GFI}(r)
        \big),
    \]
    where the Flesch--Kincaid Grade Level (FKGL) \cite{kincaid1975derivation} is
    \[
        \text{FKGL}(r) = 0.39 \times \frac{W}{S} + 11.8 \times \frac{Sy}{W} - 15.59,
    \]
    and the Gunning Fog Index (GFI) \cite{gunning} is
    \[
        \text{GFI}(r) = 0.4 \times \left( \frac{W}{S} + 100 \times \frac{C}{W} \right).
    \]
    Lower values correspond to more linguistically accessible text.
\end{definition}

In the definition above, FKGL estimates the U.S. school grade level required for comprehension, while GFI estimates the years of formal education needed to understand a text on first reading.  
These metrics are widely used in health communication research because they capture syntactic complexity, lexical difficulty, and overall patient-facing accessibility.

\begin{definition}[Affective Resonance]
    Let $r$ be a textual response.  
    Let $\sigma(r)$ be a sentiment classification function mapping $r$ to  
    $\{\text{Very Negative}, \text{Negative}, \text{Neutral}, \text{Positive}, \text{Very Positive}\}$,  
    and let $\varepsilon(r)$ be an emotion classifier that outputs a probability distribution over a set of emotions $\mathcal{E}$.  
    We define \textit{affective resonance} as
    \[
        \text{AR}(r) = \big(\sigma(r),\, \varepsilon(r)\big),
    \]
    representing the affective polarity and fine-grained emotional profile of the response.
\end{definition}

Affective resonance quantifies the emotional characteristics of a text, providing a structured approximation of empathetic tone.  
We operationalize it using two complementary affective signals:
\begin{itemize}
    \item \textit{Sentiment Classification:}  
    The \texttt{tabularisai/robust-sentiment-analysis} model \cite{vadim_borisov_2025} assigns each response to one of five sentiment classes, offering a coarse-grained measure of affective polarity.

    \item \textit{Emotion Classification:}  
    The \texttt{SamLowe/roberta-base-go\_emotions} model \cite{sam_lowe_2024} provides probabilities over 28 fine-grained emotions, including \textit{caring}, a signal associated with supportive or empathetic tone.  
    We compare emotion distributions across human and model responses via contingency analysis.
\end{itemize}

\subsection{Dataset}

To conduct our experiments, we used the MedQuAD dataset \cite{BenAbacha-TREC2017, BenAbacha-BMC-2019}, publicly available through the Hugging Face Hub\footnote{\url{https://hf.co/datasets/keivalya/MedQuad-MedicalQnADataset}}. MedQuAD contains 47{,}457 question--answer pairs extracted from 12 authoritative NIH websites, including MedlinePlus, cancer.gov, and niddk.nih.gov\footnote{\url{https://medlineplus.gov/} --- \url{https://cancer.gov/} --- \url{https://niddk.nih.gov/}}. Each entry includes metadata such as UMLS Concept Unique Identifiers (CUIs), semantic types, question focus (e.g., disease, drug, test), and topic type (e.g., treatment, side effects, diagnosis). The dataset is distributed in XML format with structured tags for questions, answers, and metadata.

For this study, we selected a subset of 16{,}400 QA pairs covering 37 question types. We excluded three sections of the original corpus: A.D.A.M. Medical Encyclopedia, MedlinePlus Drugs, and MedlinePlus Herbal Supplements. These sections account for approximately 31{,}000 entries. They follow editorial standards that differ from the core NIH sources and show substantial variation in writing style, clinical depth, and structure. Their exclusion ensures consistency in tone, source reliability, and clinical framing. This controlled subset enables fair comparisons across models.

The resulting MedQuAD subset is suitable for evaluating patient-facing language models. It primarily includes symptom-, treatment-, and diagnosis-oriented questions authored by NIH experts. Each entry contains a \textit{question} that reflects common medical concerns and a corresponding \textit{answer} derived from expert-curated institutional content. The dataset provides standardized and clinically grounded medical explanations.

In addition to MedQuAD, we used the \textit{iCliniqQAs} subset from the \textit{medical-question-answer-data} repository\footnote{\url{https://github.com/LasseRegin/medical-question-answer-data}}. This subset contains 465 real-world physician--patient question--answer pairs collected from an online medical consultation platform. The questions are written by patients and reflect spontaneous descriptions of symptoms, concerns, and contextual information. The answers are authored by licensed physicians and follow a conversational clinical style.

The inclusion of iCliniqQAs introduces naturally occurring medical dialogue into our evaluation setting. Unlike MedQuAD, which provides institutionally curated explanations, iCliniqQAs captures authentic patient concerns and real consultation dynamics. The combination of these datasets allows us to evaluate model behavior across both standardized medical communication and real-world clinical interaction scenarios.

\subsection{Models Evaluated}
\label{models}
We selected multiple large language models (LLMs) to capture different design philosophies and degrees of domain specialization. 
\texttt{Mixtral} \cite{jiang2024mixtralexperts} represents a general-purpose model trained on diverse conversational and web-scale corpora without explicit biomedical fine-tuning. 
Conversely, \texttt{Med-PaLM} \cite{singhal2022largelanguagemodelsencode} is a domain-adapted model optimized for clinical reasoning and medical question answering through supervised instruction on biomedical literature and expert-annotated data. 
This contrast enables a controlled investigation of how domain specialization influences the communicative quality of medical responses.

To further test whether observed trends generalize across distinct architectures and training paradigms, we evaluated \texttt{GPT-5}\footnote{\url{https://openai.com/index/introducing-gpt-5/}}, \texttt{Gemini 2.5 Pro} \cite{comanici2025gemini25pushingfrontier}, and \texttt{Claude Sonnet 4.5}\footnote{\url{https://www.anthropic.com/claude/sonnet}}.

\subsubsection*{MedQuAD Subset Construction}

\begin{figure}[t]
    \centering
    \includegraphics[width=1.00\linewidth]{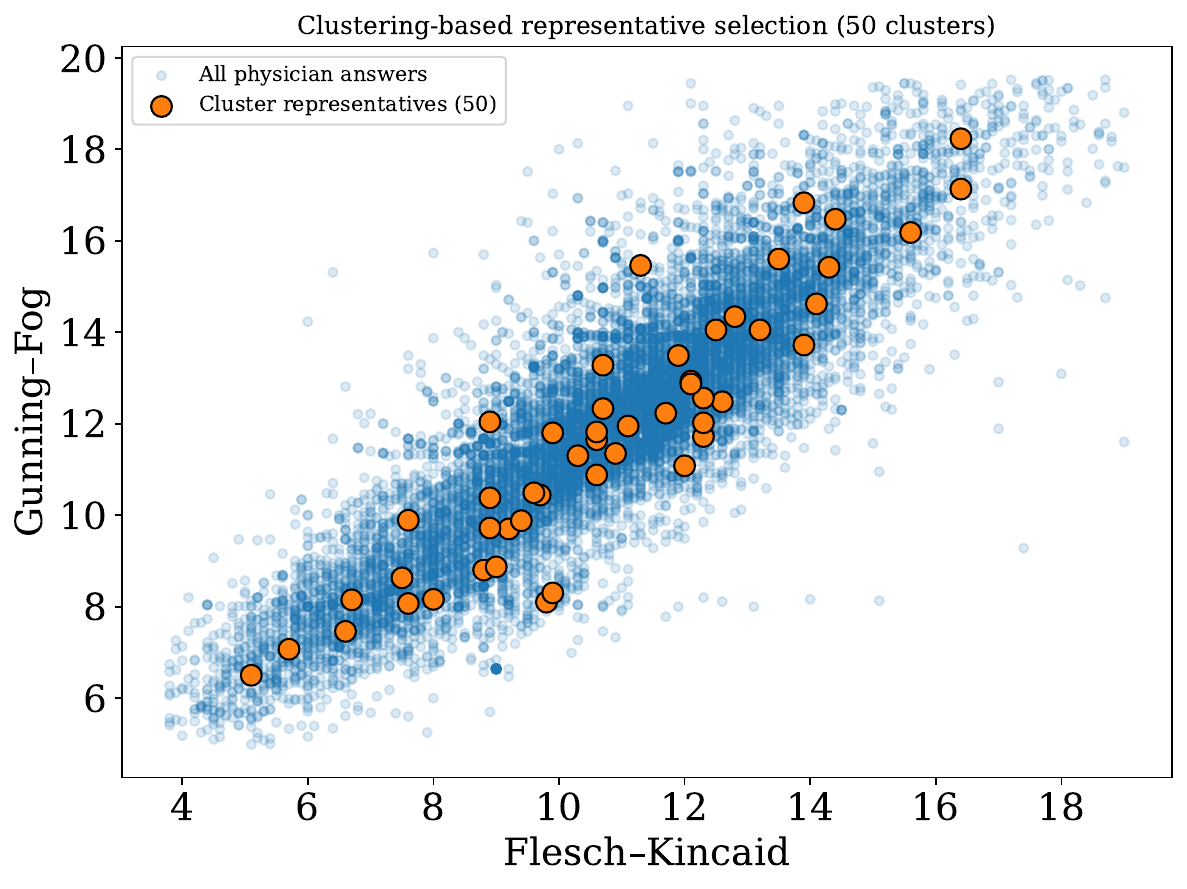}
    \caption{Readability-based selection of 50 representative MedQuAD questions after outlier removal.
    The corpus is visualized in the Gunning Fog vs. Flesch–Kincaid space.
    Extreme values were excluded using the interquartile range (IQR) criterion prior to clustering.
    The selected representatives correspond to the centroids of each $k$-means cluster ($k=50$).
    The resulting subset spans the full readability distribution of the cleaned corpus while avoiding anomalous texts that could distort clustering geometry.}
    \label{fig:clustering_representatives_medquad}
\end{figure}

Due to the high inference cost of commercial models, evaluation was conducted on a reduced subset of 50 MedQuAD questions. This subset reflects explicit budget constraints and follows a structured selection protocol designed to preserve linguistic diversity. The experiment is framed as a controlled cross-architecture comparison rather than a population-scale benchmark.

The subset was constructed through a readability-driven clustering procedure:

\begin{enumerate}
    \item \textbf{Extract linguistic features:}  
    For each question in dataset $D$, compute the Flesch--Kincaid Grade Level (FKGL), the Gunning Fog Index (GFI), lexical representativeness (cosine similarity between TF--IDF vectors and the corpus centroid), and answer length.

    \item \textbf{Remove extreme outliers:}  
    Apply interquartile range (IQR) filtering independently to FKGL and GFI.

    \item \textbf{Normalize features:}  
    Apply z-score normalization:
    \[
    z = \frac{x - \mu}{\sigma}
    \]

    \item \textbf{Cluster by linguistic properties:}  
    Perform $k$-means clustering with $k=50$ in the normalized feature space  
    $(\text{FKGL}, \text{GFI}, \text{lexical\_repr}, |\text{answer}|)$.

    \item \textbf{Select representatives:}  
    Select the question closest to each cluster centroid.

    \item \textbf{Aggregate subset:}  
    Collect selected samples into $D^{\text{MedQuAD}}_{50}$.
\end{enumerate}

Figure~\ref{fig:clustering_representatives_medquad} confirms that the retained subset spans the full readability range of the corpus.

\subsubsection*{iCliniqQAs Subset Construction}

\begin{figure}[t]
    \centering
    \includegraphics[width=1.00\linewidth]{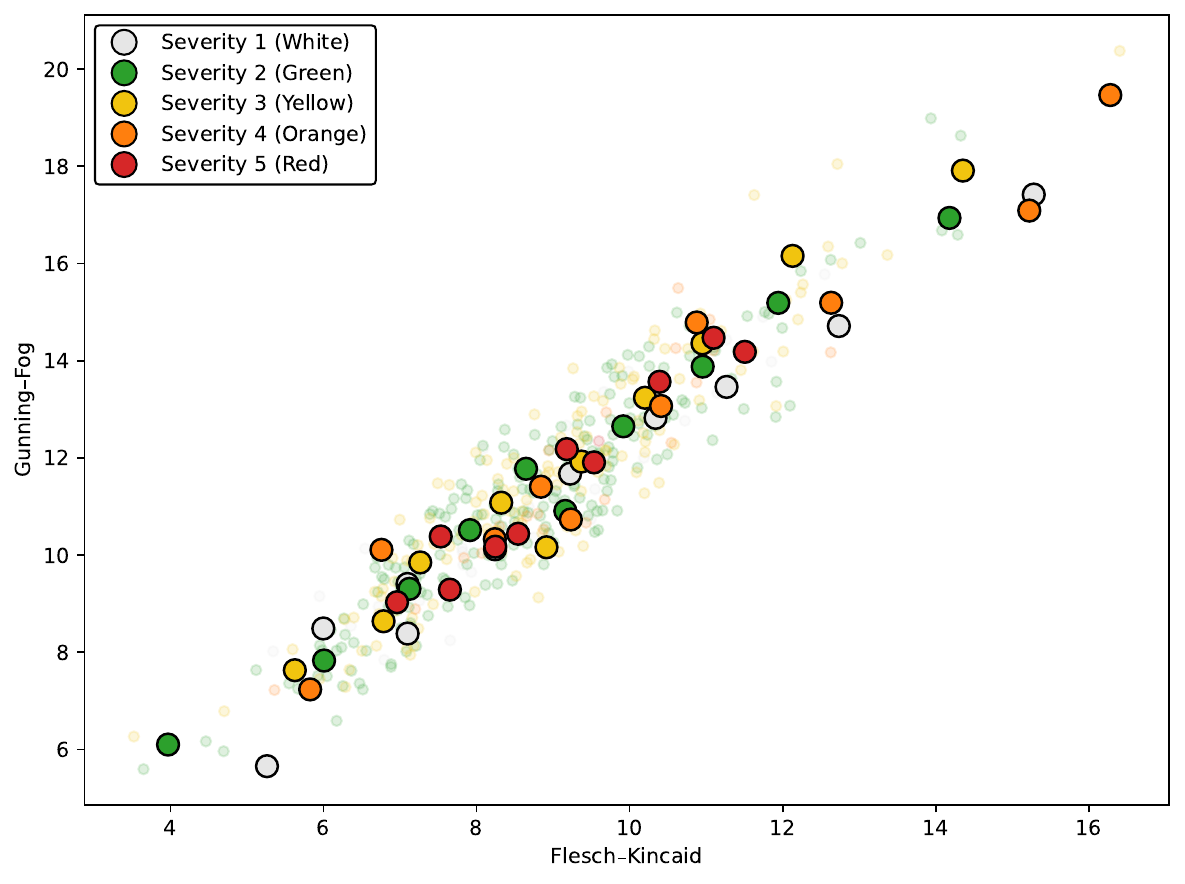}
    \caption{Severity-aware selection of 50 representative iCliniqQAs samples.
    The corpus is visualized in the Gunning Fog vs. Flesch–Kincaid space.
    Samples are stratified into five clinical severity levels (White, Green, Yellow, Orange, Red).
    Ten representatives are selected per severity class after clustering in the normalized linguistic feature space.
    The resulting subset preserves both urgency distribution and readability variability.}
    \label{fig:clustering_representatives_icliniq}
\end{figure}

For iCliniqQAs, subset construction combined linguistic stratification with clinical severity balancing. The goal was to ensure representation across urgency levels while maintaining variability in linguistic complexity.

\begin{enumerate}
    \item \textbf{Extract linguistic features:}  
    Compute FKGL, GFI, lexical representativeness, and answer length for each sample.

    \item \textbf{Normalize features:}  
    Apply z-score normalization to all linguistic features.

    \item \textbf{Severity labeling:}  
    Assign each question to one of five triage levels (White, Green, Yellow, Orange, Red).
    Labels were generated using PalMed-2 to ensure medically coherent classification.

    \item \textbf{Stratify by severity:}  
    Partition the dataset into five severity groups.

    \item \textbf{Cluster within each group:}  
    Perform clustering in the normalized linguistic feature space within each severity class.

    \item \textbf{Select balanced representatives:}  
    Select 10 samples per severity class based on centroid proximity.

    \item \textbf{Aggregate subset:}  
    Combine selected samples into $D^{\text{iCliniq}}_{50}$.
\end{enumerate}

Figure~\ref{fig:clustering_representatives_icliniq} shows that the resulting subset preserves the full range of clinical urgency while maintaining diversity in readability and lexical density.

Both subsets support controlled cross-model comparison under resource constraints. The reduced sample size limits statistical generalization but maximizes coverage across linguistic and clinical dimensions.

\subsubsection{Prompting strategies}
To evaluate how expertise and communication style affect outputs, we tested each model under three distinct response-generation settings:
\begin{itemize}
    \item \textbf{Base Prompt - Clinical Baseline Mode} (Appendix~\ref{base}): this label emphasizes that the prompt represents the model’s default, unconditioned clinical behavior, serving as a neutral reference point for all comparisons.

    \item \textbf{Empathy Prompt - Empathy-Driven Generation} (Appendix~\ref{empathy}): this name highlights that the model is explicitly instructed to generate responses with enhanced emotional awareness and patient-centered tone, framing the prompt as an affective alignment strategy rather than a mere style change.

    \item \textbf{Rephrase Prompt - AI-Assisted Clinical Editing} (Appendix~\ref{rephrase}): this formulation clarifies that the model operates as a collaborative editor, reframing its role from content generator to clinical communication enhancer\cite{DBLP:journals/npjdm/MandalWSSMRSMN25}.
\end{itemize}

Together, these three configurations isolate complementary aspects of communicative alignment: the Base Prompt captures factual generation, the Empathy Prompt evaluates stylistic modulation during autonomous generation, and the Rephrase Prompt measures the model’s capacity to enhance existing human-authored content through collaborative refinement.

\section{Experiments and Results}\label{sec4}

In this section, we present the experimental setup, describe the evaluation procedures, and report the results for each research question (RQ). Before addressing the RQs individually, we first evaluate semantic fidelity across systems to establish a baseline understanding of how closely LLM-generated responses align with physician-written content.

\subsection{Experimental Setup} 

Our experimental setup was structured in two main phases.

\subsubsection{Phase 1 – Model Comparison}
In the first phase of the study, we generated four responses for each question in the MedQuAD and iCliniqQAs subsets by combining two language models - \texttt{Mixtral} and \texttt{Med-PaLM 2} - with two prompting strategies: the \textit{Base Prompt} and the \textit{Empathy Prompt}. This setup yielded four distinct outputs, representing general-purpose and medical-domain generations under both standard and empathy-enhanced conditions. Each output was then systematically compared with the physician-authored reference answer, resulting in a structured five-way evaluation for every question.


To further examine the generalizability of the observed trends, we extended the evaluation to additional architectures-\texttt{GPT-5}, \texttt{Gemini 2.5 Pro}, and \texttt{Claude Sonnet 4.5}-using a representative subset of 50 questions selected through a readability-based clustering procedure. For \texttt{Gemini 2.5 Pro}, however, a complete evaluation was not feasible: the model frequently produced limited or incomplete answers when contextual information was insufficient, underscoring its reliance on external context to generate medically grounded responses. This behavior also reflected an ethical safeguard, as the model tended to refrain from producing potentially unreliable clinical information in the absence of adequate medical context.

\subsubsection{Phase 2 – Physician Answer Rephrase}
To investigate whether large language models can assist or refine physician-authored responses, we employed the same set of models as in the previous experiments: \texttt{Mixtral}, \texttt{Med-PaLM 2}, \texttt{GPT-5}, \texttt{Gemini 2.5 Pro}, and \texttt{Claude Sonnet 4.5}, to rewrite each original medical answer.
The resulting generations are denoted as \texttt{Model\_Rephrase}, following a consistent naming convention across models (e.g., \texttt{Mixtral\_Rephrase}, \texttt{Med-PaLM\_Rephrase}, etc.). Using a dedicated rewriting prompt, each model was instructed to enhance the emotional tone, clarity, and accessibility of the physician’s message while preserving its medical accuracy and factual consistency.

This phase simulated a human–AI co-authoring process distinct from the \textit{Base Prompt} and \textit{Empathy Prompt (Empathy Prompt)} configurations used in Phase 1, emphasizing collaborative refinement rather than autonomous response generation.
All outputs were generated with low-temperature sampling ($temperature = 0.1$) to limit stochastic variation and promote consistency across runs.

\subsection{Preliminary Evaluation – Semantic Fidelity}
\begin{figure}[t]
    \centering
    \includegraphics[width=1.0\textwidth]{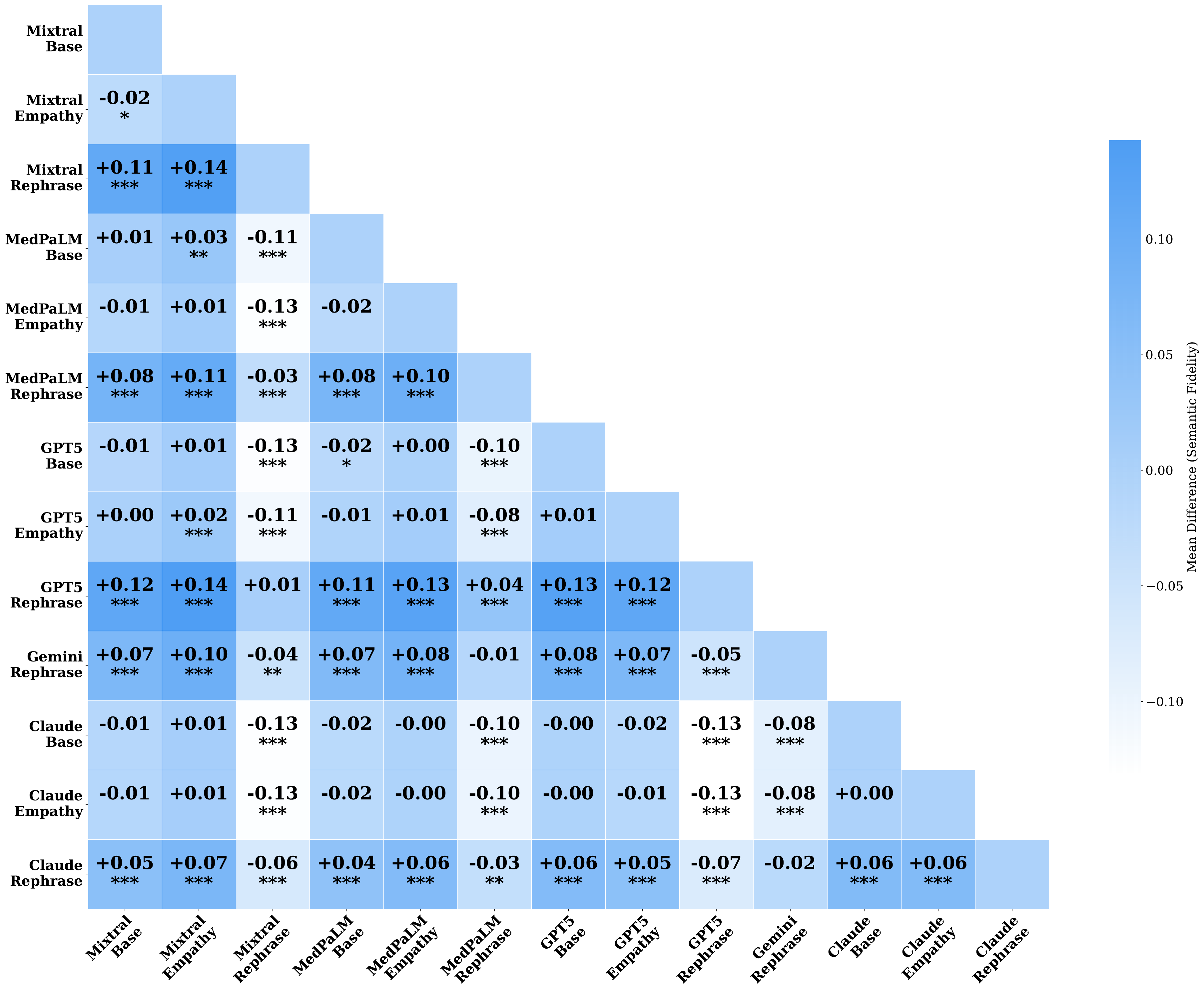}
    \caption{Pairwise comparison of language models in terms of semantic fidelity on the MedQuAD dataset. Each cell reports the mean difference in semantic fidelity between model pairs (Model $i - j$), where positive values indicate higher similarity to the medical reference for the model reported on the row. Color intensity encodes the magnitude of the difference, while statistical significance after FDR correction is indicated by asterisks ($^* p < 0.05$, $^{**} p < 0.01$, $^{***} p < 0.001$).}

    \label{fig:heatmappvalue}
\end{figure}

\begin{figure}[t]
    \centering
    \includegraphics[width=1.0\textwidth]{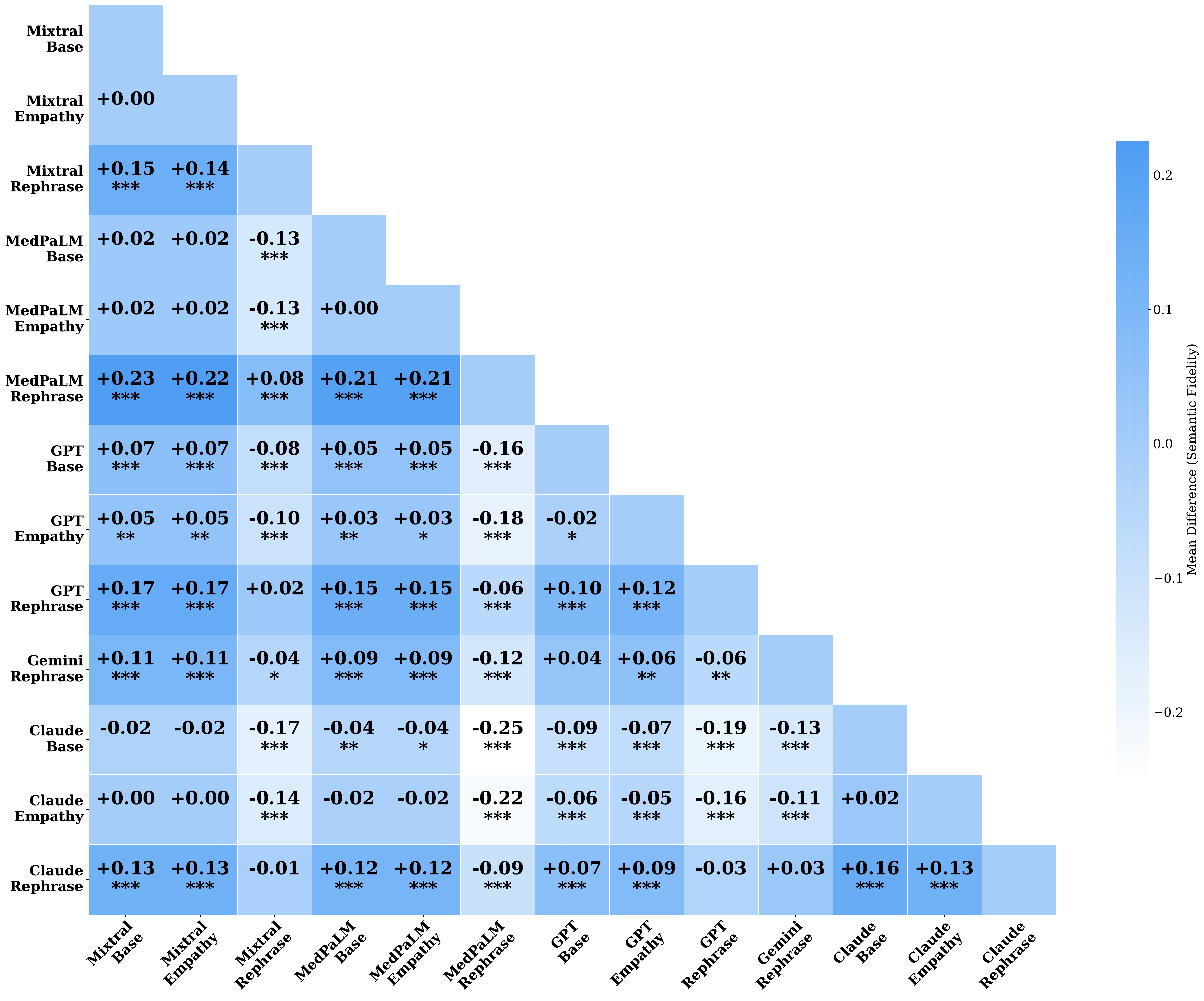}
    \caption{Pairwise comparison of language models in terms of semantic fidelity on the iCliniqQAs dataset. 
    Each cell reports the mean difference in semantic fidelity between model pairs (Model $i - j$), 
    where positive values indicate higher similarity to the medical reference for the model reported on the row. 
    Color intensity encodes the magnitude of the difference, while statistical significance after FDR correction 
    is indicated by asterisks ($^* p < 0.05$, $^{**} p < 0.01$, $^{***} p < 0.001$).}
    
    \label{fig:heatmappvalue_icliniq}
\end{figure}

Semantic fidelity was evaluated as a prerequisite validation step to verify that LLM-generated responses are semantically aligned with physician-authored answers before conducting downstream analyses.
Cosine similarity was computed between sentence embeddings obtained with the \texttt{BioBERT-mnli-snli-scinli-scitail-mednli-stsb}\footnote{\url{https://huggingface.co/pritamdeka/BioBERT-mnli-snli-scinli-scitail-mednli-stsb}} model \cite{deka2022evidence}.
Descriptive statistics for each configuration are reported in Figure~\ref{fig:heatmappvalue} and Figure~\ref{fig:heatmappvalue_icliniq}.

All evaluated systems exhibit strong conceptual alignment with physician responses across both datasets.
Average cosine similarity values are consistently above $0.78$ in the first dataset and above $0.75$ in the iCliniqQAs dataset.

\begin{figure}[t]
    \centering
    \includegraphics[width=\textwidth]{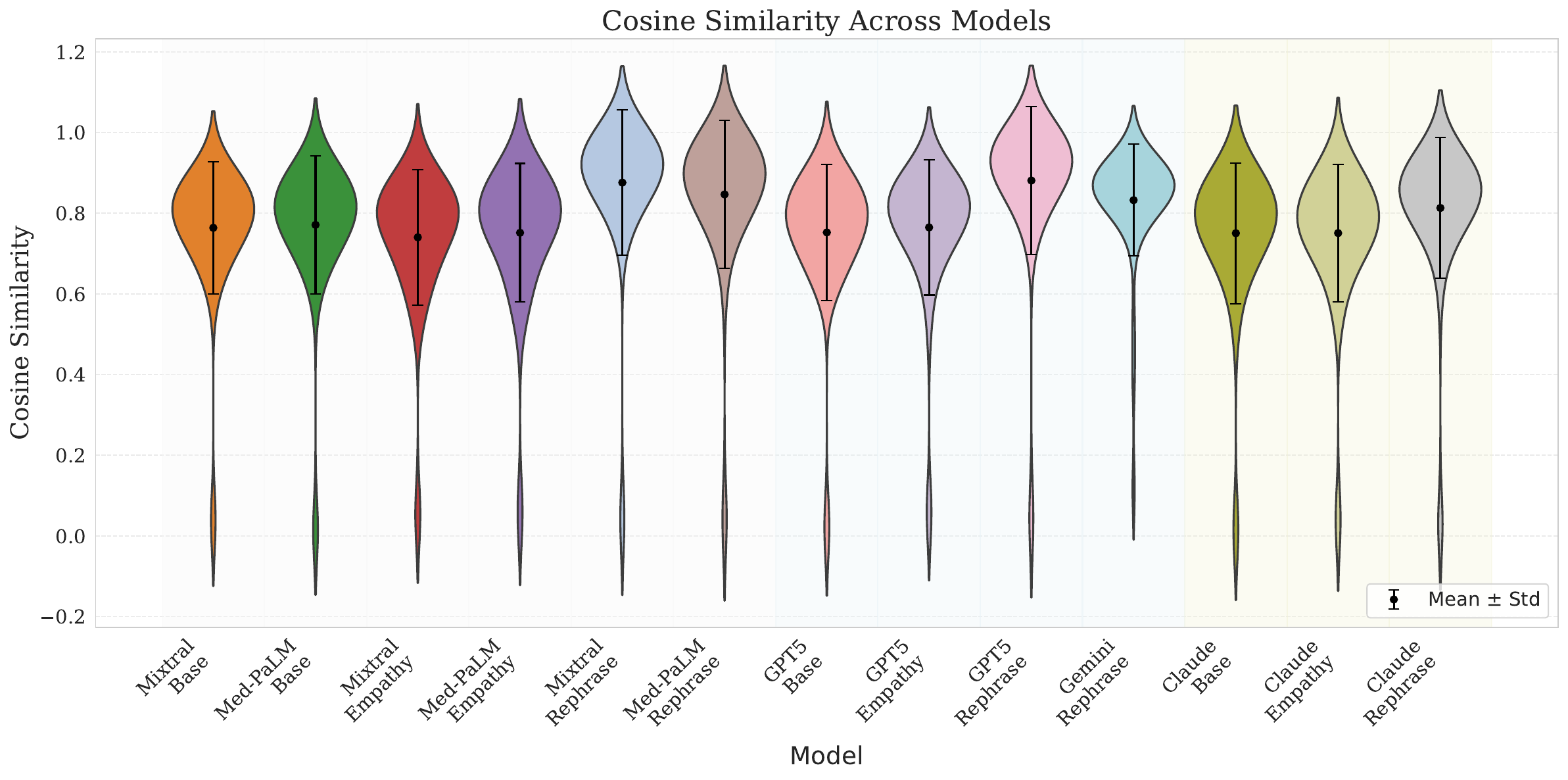}
    \caption{Cosine similarity between physician-written and model-generated answers on the MedQuAD dataset.
    Higher values reflect closer semantic alignment. 
    Gemini 2.5 Pro appears only in the Rephrase configuration because, in our experiments, 
    the model frequently refused to generate Base or Empathy responses without sufficient 
    clinical context, exhibiting strong safety guardrails similar to those observed in 
    Claude\_Base (left in the comparison as an explicit example of this behaviour).}
    
    \label{fig:similarity_violinplot}
\end{figure}

\begin{figure}[t]
    \centering
    \includegraphics[width=\textwidth]{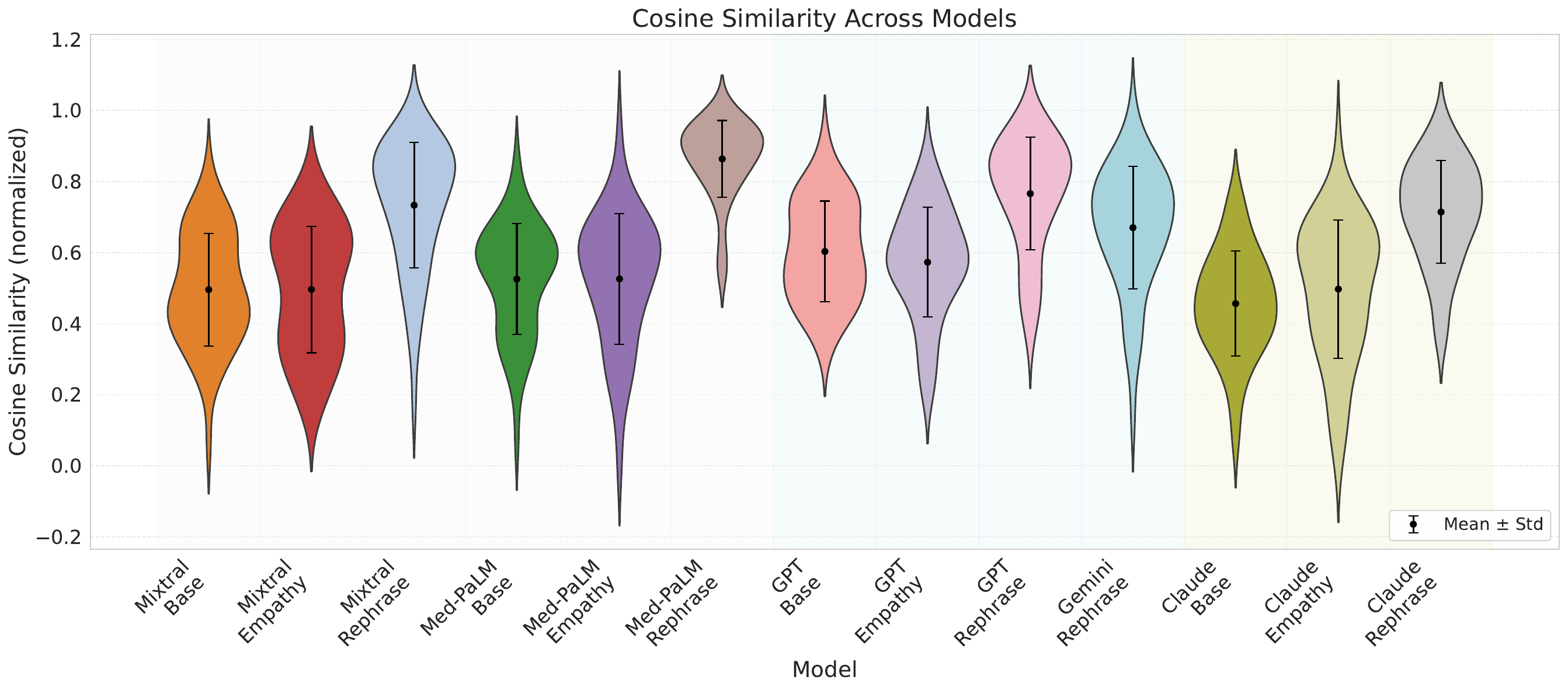}
    \caption{Cosine similarity between physician-written and model-generated answers on the iCliniqQAs dataset.
    Higher values reflect closer semantic alignment. 
    Gemini 2.5 Pro appears only in the Rephrase configuration because, in our experiments, 
    the model frequently refused to generate Base or Empathy responses without sufficient 
    clinical context, exhibiting strong safety guardrails similar to those observed in 
    Claude\_Base (left in the comparison as an explicit example of this behaviour).}
    
    \label{fig:similarity_violinplot_icliniq}
\end{figure}

In the first dataset as we can see in figure \ref{fig:similarity_violinplot}, the highest semantic fidelity is achieved by \textit{GPT5\_Rephrase} ($\mu = 0.92$), followed by \textit{Mixtral\_Rephrase} ($\mu = 0.91$) and \textit{MedPaLM\_Rephrase} ($\mu = 0.89$), while \textit{Gemini\_Rephrase} and \textit{Claude\_Rephrase} reach $\mu = 0.87$ and $\mu = 0.85$, respectively.
In contrast, on the iCliniqQAs dataset (Figure~\ref{fig:similarity_violinplot_icliniq}), the highest semantic fidelity is achieved by \textit{MedPaLM\_Rephrase} ($\mu = 0.93$), followed by \textit{GPT\_Rephrase} ($\mu = 0.91$) and \textit{Mixtral\_Rephrase} ($\mu = 0.89$), with \textit{Gemini\_Rephrase} ($\mu = 0.86$) and \textit{Claude\_Rephrase} ($\mu = 0.84$) showing comparatively lower performance.
Notably, the separation between domain-specialized and general-purpose models is more pronounced in iCliniqQAs, where \textit{MedPaLM\_Rephrase} consistently outperforms all other configurations.

Among baseline architectures, performance remains tightly clustered in both datasets.
In the first dataset, \textit{MedPaLM\_Base} ($\mu = 0.82$), \textit{Mixtral\_Base} ($\mu = 0.80$), and \textit{GPT5\_Base} ($\mu = 0.79$) show closely matched alignment.
In iCliniqQAs, \textit{MedPaLM\_Base} ($\mu = 0.80$) and \textit{Mixtral\_Base} ($\mu = 0.78$) remain comparable, while \textit{GPT\_Base} ($\mu = 0.85$) exhibits slightly higher raw similarity but does not consistently match the gains observed in domain-adapted rephrasing configurations.

Prompt-based variants yield comparable distributions across both datasets.
In the first dataset, \textit{GPT5\_Empathy} ($\mu = 0.81$), \textit{MedPaLM\_Empathy} ($\mu = 0.80$), \textit{Claude\_Empathy} ($\mu = 0.80$), and \textit{Mixtral\_Empathy} ($\mu = 0.78$) remain aligned with baseline levels.
Similarly, in iCliniqQAs, \textit{GPT\_Empathy} ($\mu = 0.83$), \textit{MedPaLM\_Empathy} ($\mu = 0.80$), \textit{Claude\_Empathy} ($\mu = 0.78$), and \textit{Mixtral\_Empathy} ($\mu = 0.78$) show limited deviation from their respective base configurations, confirming that empathy prompting alone does not substantially increase semantic fidelity.

Statistical significance between model configurations was assessed via two-sided paired $t$-tests with False Discovery Rate (FDR) correction using the Benjamini--Hochberg procedure~\cite{benjamini1995controlling}.
Each statistical population corresponds to the distribution of cosine similarity scores produced by a model across all evaluated questions.
Let $m_1$ and $m_2$ denote two distinct model configurations and $\mu_m$ the associated mean similarity.
For each pairwise comparison, the null hypothesis is defined as $H_0: \mu_{m_1} = \mu_{m_2}$.

No statistically significant difference is observed between \textit{Mixtral\_Base} and \textit{MedPaLM\_Base} in either dataset, confirming comparable semantic fidelity at baseline.
Rephrasing configurations introduce systematic improvements across both datasets; however, the effect is particularly pronounced in the iCliniqQAs dataset, where \textit{MedPaLM\_Rephrase} achieves statistically significant improvements over both its baseline and empathy variants as well as over multiple general-purpose counterparts ($p < 0.01$, FDR-corrected).
The full matrices of mean differences and FDR-adjusted $p$-values are depicted in Figure~\ref{fig:heatmappvalue} and Figure~\ref{fig:heatmappvalue_icliniq}, highlighting statistically significant contrasts across multiple model pairs, especially those involving domain-specialized rephrasing configurations.

\begin{takeaway}
Rewriting consistently yields the highest semantic fidelity across both datasets.
In MedQuAD, \textit{GPT5\_Rephrase} achieves the strongest alignment with physician-authored answers ($\mu = 0.92$), 
while in iCliniqQAs the best performance is obtained by \textit{MedPaLM\_Rephrase} ($\mu = 0.93$). 
Across architectures, rephrase configurations systematically outperform both baseline and empathy-prompted variants, 
confirming collaborative rewriting as the most effective strategy for maximizing conceptual overlap with clinical experts.
\end{takeaway}

\subsection{RQ1 – Empathy and Sentiment Analyses}
This research question evaluates whether LLMs can produce responses that match physician-authored texts in terms of emotional attunement. To this end, we conduct a two-step evaluation using both general sentiment classification and fine-grained emotion detection, using the models described in the \textit{Background} section.

\begin{hyp}
\textit{
Let $R_{\text{LLM}}$ be a response generated by an LLM, and let $R_{\text{Phys}}$ be a physician-authored response. Let $E(\cdot)$ denote the affective resonance function introduced in Section~3.1, which captures both sentiment polarity and fine-grained emotional expression. 
We hypothesize that LLM-generated responses exhibit comparable affective resonance to physician-authored ones, i.e.,
\[
\mathbb{E}[E(R_{\text{LLM}})] = \mathbb{E}[E(R_{\text{Phys}})].
\]
}
\end{hyp}

\subsubsection{Sentiment Distribution}
\begin{figure}[htbp]
    \centering
    \includegraphics[width=\textwidth]{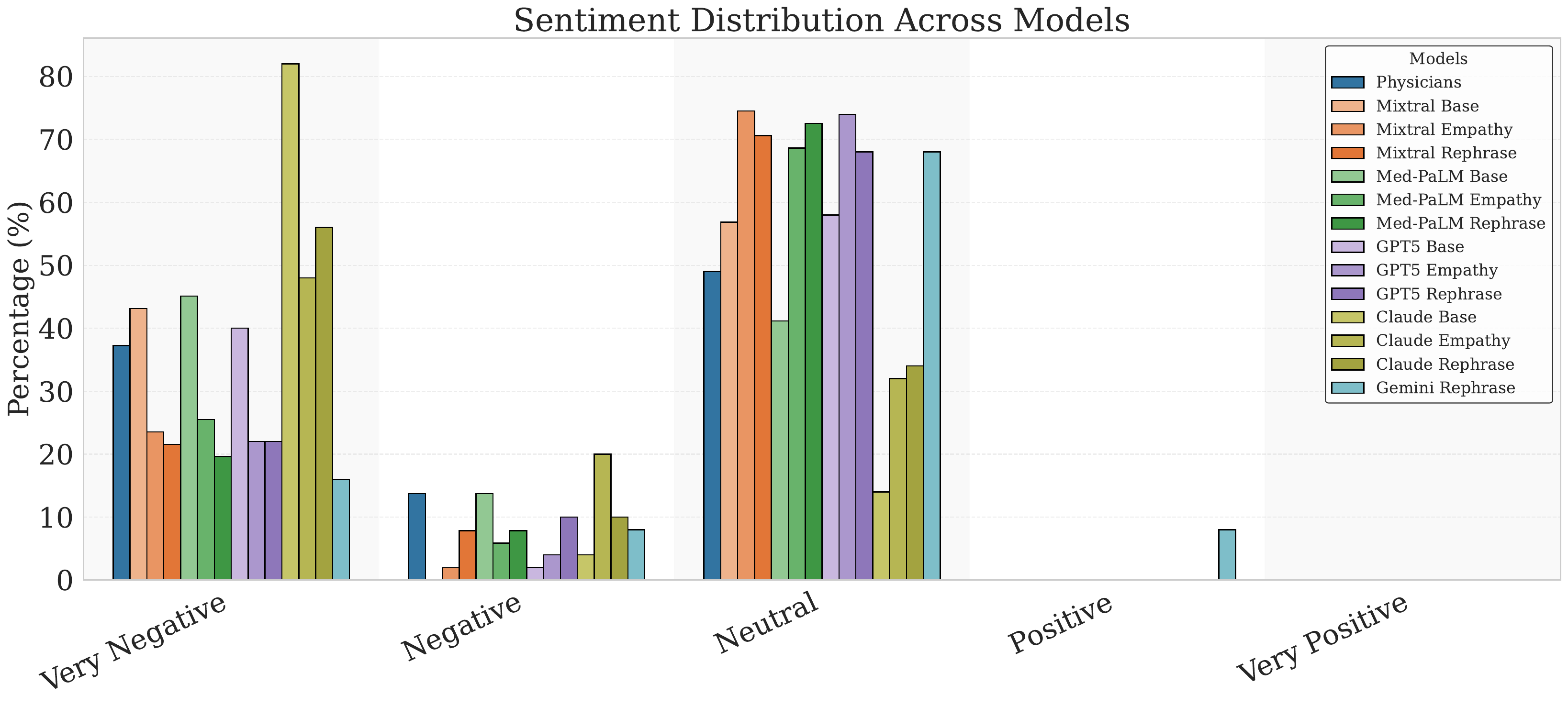}
    \caption{Distribution of sentiment expressed by models on the MedQuAD dataset.}
    \label{fig:sentiment_distribution}
\end{figure}

\begin{figure}[htbp]
    \centering
    \includegraphics[width=\textwidth]{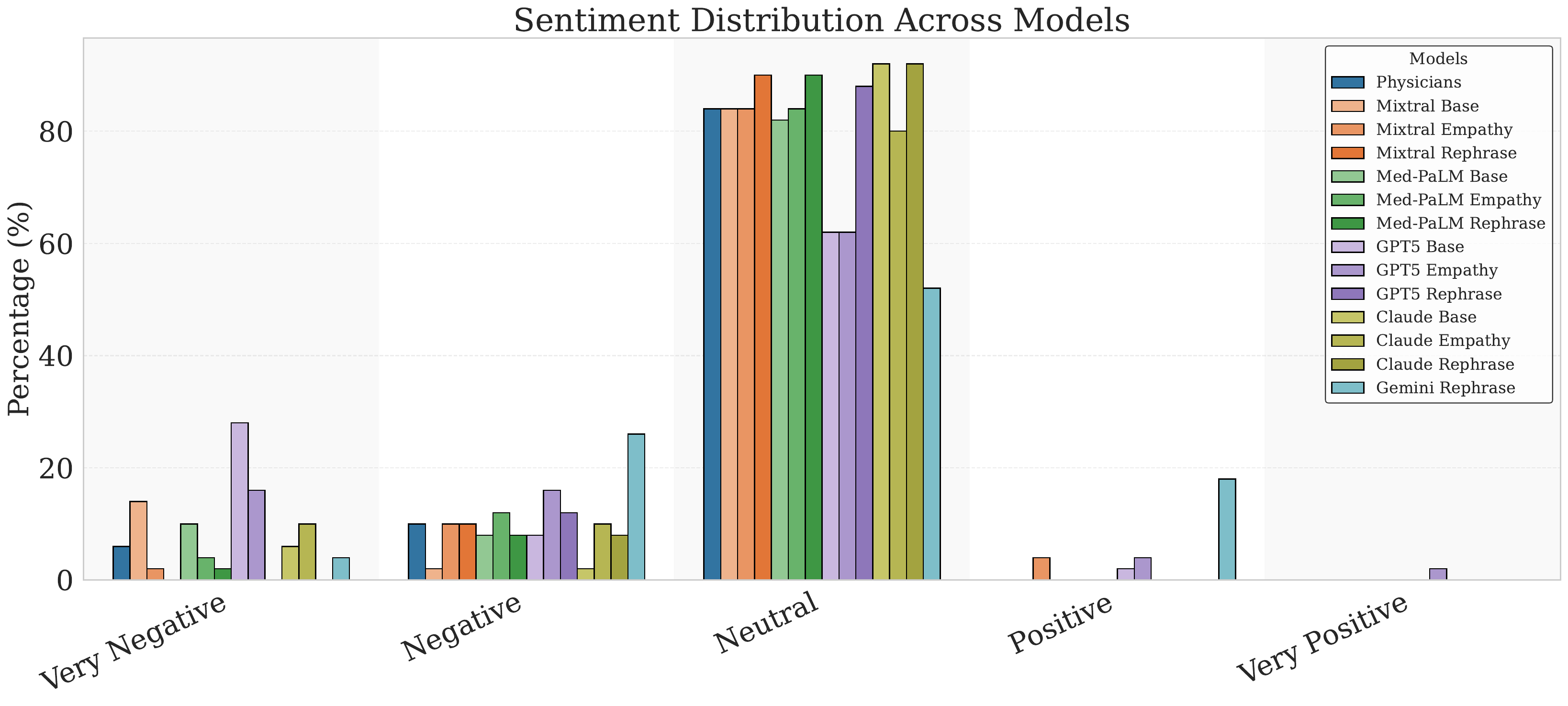}
    \caption{Distribution of sentiment expressed by models on the iCliniqQAs dataset.}
    \label{fig:sentiment_distribution_icliniq}
\end{figure}

We categorized each response into one of 5 sentiment classes: \textit{Very Negative}, 
\textit{Negative}, \textit{Neutral}, \textit{Positive}, and \textit{Very Positive}.
As shown in Figures~\ref{fig:sentiment_distribution} and \ref{fig:sentiment_distribution_icliniq}, physicians' responses predominantly 
fall into the \textit{Neutral} category.

From Table~\ref{tab:sentiment_distribution_transposed}, physician answers in the MedQuAD 
dataset concentrate predominantly in the \textit{Neutral} category (49.02\%), with a 
substantial proportion of \textit{Very Negative} responses (37.25\%) and virtually no 
positive affect. As illustrated in Figure~\ref{fig:sentiment_distribution}, this 
distribution reflects a clinically restrained tone typical of institutional medical 
communication. In contrast, baseline LLM configurations on MedQuAD tend to amplify 
polarity. Both \textit{Mixtral} and \textit{Med-PaLM} increase the proportion of 
\textit{Very Negative} responses (43.14\% and 45.10\%, respectively), indicating a 
sharper affective framing than physicians. Prompt-based and rephrased configurations 
mitigate this effect, systematically shifting outputs toward higher \textit{Neutral} 
rates and reducing extreme negativity. Notably, \textit{Gemini\_Rephrase} is the only 
configuration exhibiting a non-negligible proportion of \textit{Positive} sentiment 
(8.0\%), suggesting a mild but distinct tendency toward affective reinforcement absent 
from physician-authored texts.

A different pattern emerges in the second dataset (iCliniqQAs), where physician 
responses are even more strongly dominated by \textit{Neutral} sentiment (84.0\%) and 
contain markedly lower levels of \textit{Very Negative} content (6.0\%). This reflects 
the conversational and patient-facing nature of the dataset, in which clinicians adopt 
a less confrontational and more stabilizing tone. In this setting, baseline models do 
not systematically amplify extreme negativity as observed in MedQuAD; instead, they 
display greater variability in the distribution of \textit{Negative} and \textit{Neutral} 
responses. Rephrasing strategies generally increase \textit{Neutral} proportions (e.g., 
up to 90--92\% in several configurations), further aligning outputs with physician 
affective restraint. However, \textit{Gemini\_Rephrase} again stands out, exhibiting a 
substantially higher proportion of \textit{Positive} sentiment (18.0\%), a level not 
observed in physician responses in either dataset.

\begin{figure*}[t]
  \centering
  \includegraphics[width=1.0\textwidth]{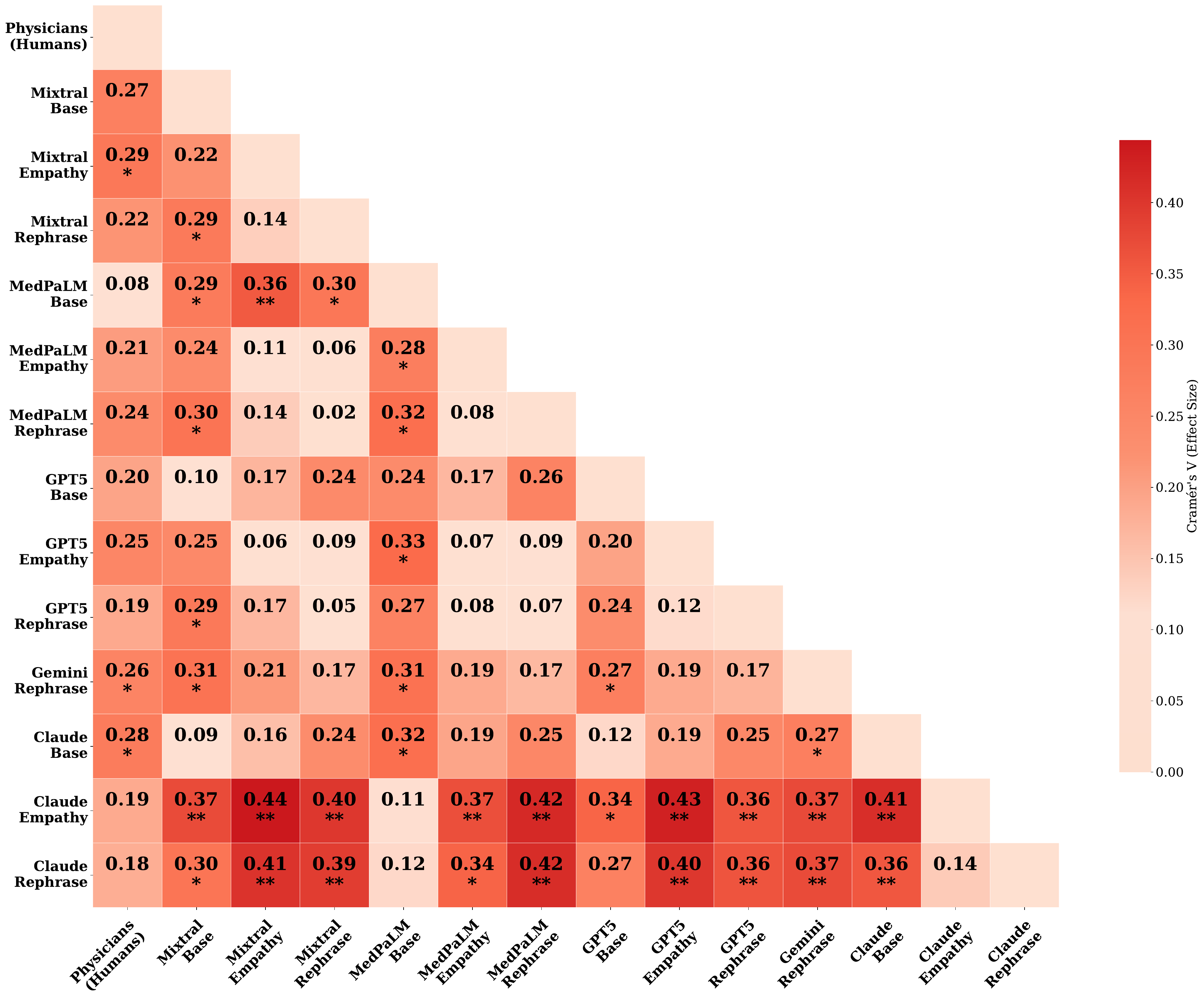}
  \caption{
  Pairwise comparison of sentiment distributions across systems on the MedQuAD dataset.
  Cells report Cramér’s V effect size for each model pair; darker color indicates larger divergence.
  Asterisks denote FDR-corrected significance (* $p<0.05$, ** $p<0.01$, *** $p<0.001$).}
  \label{fig:sentiment_heatmap}
\end{figure*}

\begin{figure*}[t]
  \centering
  \includegraphics[width=1.0\textwidth]{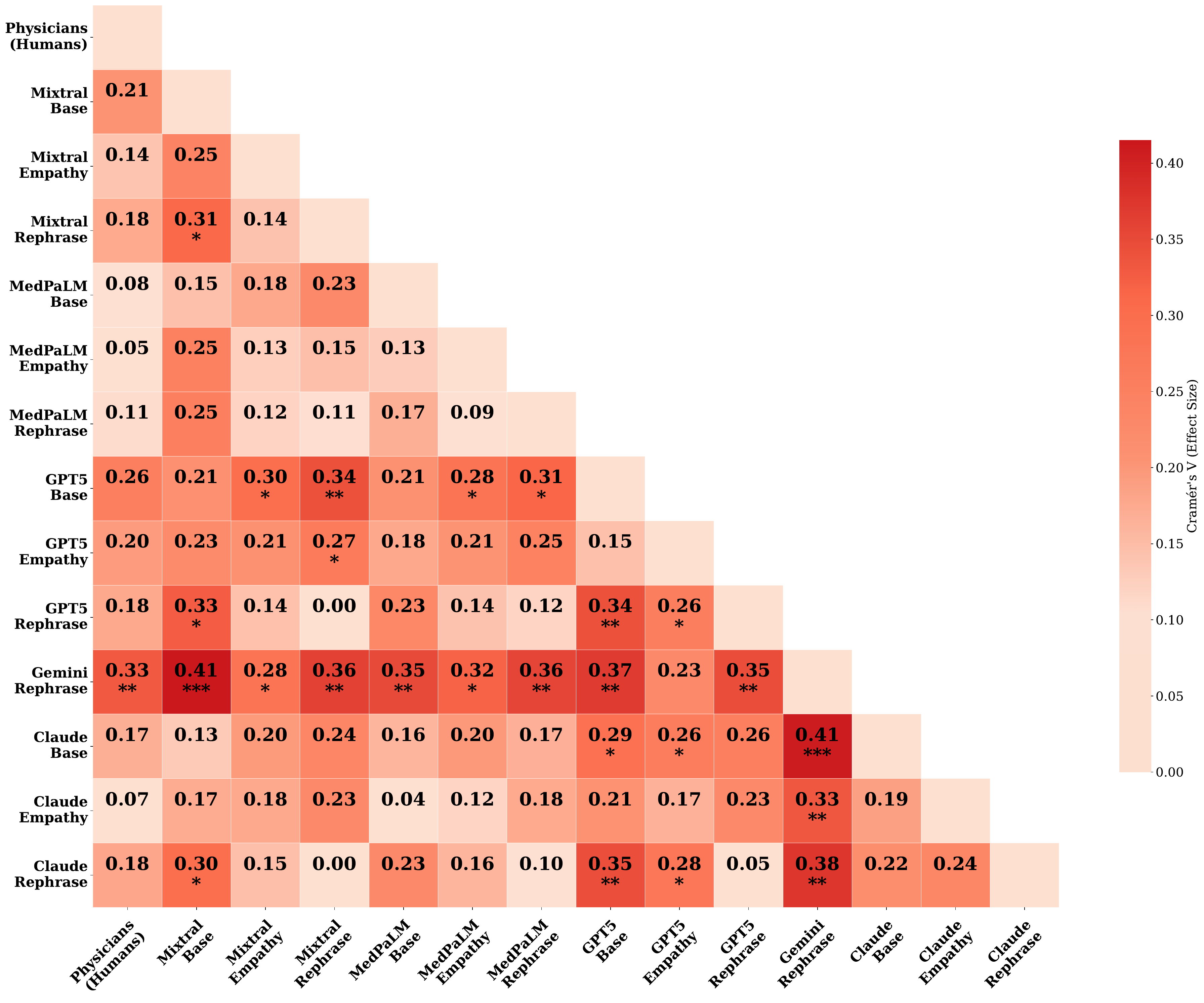}
  \caption{
  Pairwise comparison of sentiment distributions across systems on the iCliniqQAs dataset.
  Cells report Cramér’s V effect size for each model pair; darker color indicates larger divergence.
  Asterisks denote FDR-corrected significance (* $p<0.05$, ** $p<0.01$, *** $p<0.001$).}
  \label{fig:sentiment_heatmap1}
\end{figure*}

Pairwise chi-square analyses with Benjamini--Hochberg correction confirm that these 
deviations are not uniform across systems (Figures~\ref{fig:sentiment_heatmap} 
and~\ref{fig:sentiment_heatmap1}). In the 
MedQuAD setting, \textit{Claude (Base)} exhibits the strongest divergence from 
physicians ($V = 0.45$, $p < 0.001$). This result, however, does not reflect 
polarity amplification of the same kind observed in \textit{Mixtral} and 
\textit{Med-PaLM}: as noted in Section~\ref{sec4}, Claude Base 
frequently produced cautious, hedged responses in the absence of sufficient clinical 
context-analogous to the safety-driven refusals observed in Gemini. These outputs 
were nonetheless classified by the sentiment model, yielding a disproportionately 
high \textit{Very Negative} rate (82.0\%) that reflects classifier sensitivity to 
evasive or uncertainty-laden language rather than affectively charged clinical 
content. By contrast, \textit{Med-PaLM\_Base} remains closest to physician 
distributions ($V = 0.08$, $p > 0.05$), representing the only baseline configuration 
whose sentiment profile is not statistically distinguishable from that of 
physician-authored responses. In the iCliniqQAs dataset, effect sizes are generally 
more moderate, indicating closer overall alignment with human-authored sentiment 
patterns, though statistically significant differences persist for selected 
configurations.

\begin{table*}
\centering
\caption{Percentage distribution of sentiment labels per system on MedQuAD dataset. Arrows indicate comparison to Doctor: \textcolor{red}{$\uparrow$} = higher, \textcolor{blue}{$\downarrow$} = lower, \textcolor{gray}{-} = similar.}
\label{tab:sentiment_distribution_transposed}
\resizebox{0.9\textwidth}{!}{
\begin{tabular}{lccccc}
\toprule
\textbf{System} & \textbf{Very Negative (\%)} & \textbf{Negative (\%)} & \textbf{Neutral (\%)} & \textbf{Positive (\%)} & \textbf{Very Positive (\%)} \\
\midrule
\textit{Physician Answer}    
& \textit{37.25} 
& \textit{13.73} 
& \textit{49.02} 
& \textit{0.00} 
& \textit{0.00} \\
\midrule

Mixtral 
& 43.14 ({\color{red}↑}) 
& 0.00 ({\color{blue}↓}) 
& 56.86 ({\color{red}↑}) 
& 0.00 ({\color{gray}---}) 
& 0.00 ({\color{gray}---}) \\

Med-PaLM 
& 45.10 ({\color{red}↑}) 
& 13.73 ({\color{gray}---}) 
& 41.18 ({\color{blue}↓}) 
& 0.00 ({\color{gray}---}) 
& 0.00 ({\color{gray}---}) \\

Mixtral\_Empathy Prompt
& 23.53 ({\color{blue}↓})
& 1.96 ({\color{blue}↓})
& 74.51 ({\color{red}↑})
& 0.00 ({\color{gray}---})
& 0.00 ({\color{gray}---}) \\

Med-PaLM\_Empathy Prompt
& 25.49 ({\color{blue}↓})
& 5.88 ({\color{blue}↓})
& 68.63 ({\color{red}↑})
& 0.00 ({\color{gray}---})
& 0.00 ({\color{gray}---}) \\

Mixtral\_Rephrase
& 21.57 ({\color{blue}↓})
& 7.84 ({\color{blue}↓})
& 70.59 ({\color{red}↑})
& 0.00 ({\color{gray}---})
& 0.00 ({\color{gray}---}) \\

Med-PaLM\_Rephrase
& 19.61 ({\color{blue}↓})
& 7.84 ({\color{blue}↓})
& 72.55 ({\color{red}↑})
& 0.00 ({\color{gray}---})
& 0.00 ({\color{gray}---}) \\

GPT-5 (BASE)
& 40.00 ({\color{red}↑})
& 2.00 ({\color{blue}↓})
& 58.00 ({\color{red}↑})
& 0.00 ({\color{gray}---})
& 0.00 ({\color{gray}---}) \\

GPT-5\_Empathy Prompt
& 22.00 ({\color{blue}↓})
& 4.00 ({\color{blue}↓})
& 74.00 ({\color{red}↑})
& 0.00 ({\color{gray}---})
& 0.00 ({\color{gray}---}) \\

GPT-5\_Rephrase
& 22.00 ({\color{blue}↓})
& 10.00 ({\color{blue}↓})
& 68.00 ({\color{red}↑})
& 0.00 ({\color{gray}---})
& 0.00 ({\color{gray}---}) \\

Gemini\_Rephrase
& 16.00 ({\color{blue}↓})
& 8.00 ({\color{blue}↓})
& 68.00 ({\color{red}↑})
& 8.00 ({\color{red}↑})
& 0.00 ({\color{gray}---}) \\

Claude (BASE)
& 82.00 ({\color{red}↑})
& 4.00 ({\color{blue}↓})
& 14.00 ({\color{blue}↓})
& 0.00 ({\color{gray}---})
& 0.00 ({\color{gray}---}) \\

Claude\_Empathy Prompt
& 48.00 ({\color{red}↑})
& 20.00 ({\color{red}↑})
& 32.00 ({\color{blue}↓})
& 0.00 ({\color{gray}---})
& 0.00 ({\color{gray}---}) \\

Claude\_Rephrase
& 56.00 ({\color{red}↑})
& 10.00 ({\color{blue}↓})
& 34.00 ({\color{blue}↓})
& 0.00 ({\color{gray}---})
& 0.00 ({\color{gray}---}) \\

\bottomrule
\end{tabular}}
\end{table*}

\begin{table*}
\centering
\caption{Percentage distribution of sentiment labels per system on the iCliniqQAs dataset. Arrows indicate comparison to Physician: \textcolor{red}{$\uparrow$} = higher, \textcolor{blue}{$\downarrow$} = lower, \textcolor{gray}{-} = similar.}
\label{tab:sentiment_distribution_second_dataset}
\resizebox{0.9\textwidth}{!}{
\begin{tabular}{lccccc}
\toprule
\textbf{System} & \textbf{Very Negative (\%)} & \textbf{Negative (\%)} & \textbf{Neutral (\%)} & \textbf{Positive (\%)} & \textbf{Very Positive (\%)} \\
\midrule

\textit{Physician Answer}    
& \textit{6.0} 
& \textit{10.0} 
& \textit{84.0} 
& \textit{0.0} 
& \textit{0.0} \\
\midrule

Mixtral
& 14.0 ({\color{red}↑}) 
& 2.0 ({\color{blue}↓}) 
& 84.0 ({\color{gray}---}) 
& 0.0 ({\color{gray}---}) 
& 0.0 ({\color{gray}---}) \\

Med-PaLM
& 10.0 ({\color{red}↑})
& 8.0 ({\color{blue}↓})
& 82.0 ({\color{blue}↓})
& 0.0 ({\color{gray}---})
& 0.0 ({\color{gray}---}) \\

Mixtral\_Empathy Prompt
& 2.0 ({\color{blue}↓})
& 10.0 ({\color{gray}---})
& 84.0 ({\color{gray}---})
& 4.0 ({\color{red}↑})
& 0.0 ({\color{gray}---}) \\

Med-PaLM\_Empathy Prompt
& 4.0 ({\color{blue}↓})
& 12.0 ({\color{red}↑})
& 84.0 ({\color{gray}---})
& 0.0 ({\color{gray}---})
& 0.0 ({\color{gray}---}) \\

Mixtral\_Rephrase
& 0.0 ({\color{blue}↓})
& 10.0 ({\color{gray}---})
& 90.0 ({\color{red}↑})
& 0.0 ({\color{gray}---})
& 0.0 ({\color{gray}---}) \\

Med-PaLM\_Rephrase
& 2.0 ({\color{blue}↓})
& 8.0 ({\color{blue}↓})
& 90.0 ({\color{red}↑})
& 0.0 ({\color{gray}---})
& 0.0 ({\color{gray}---}) \\

GPT-5 (BASE)
& 28.0 ({\color{red}↑})
& 8.0 ({\color{blue}↓})
& 62.0 ({\color{blue}↓})
& 2.0 ({\color{red}↑})
& 0.0 ({\color{gray}---}) \\

GPT-5\_Empathy Prompt
& 16.0 ({\color{red}↑})
& 16.0 ({\color{red}↑})
& 62.0 ({\color{blue}↓})
& 4.0 ({\color{red}↑})
& 2.0 ({\color{red}↑}) \\

GPT-5\_Rephrase
& 0.0 ({\color{blue}↓})
& 12.0 ({\color{red}↑})
& 88.0 ({\color{red}↑})
& 0.0 ({\color{gray}---})
& 0.0 ({\color{gray}---}) \\

Gemini\_Rephrase
& 4.0 ({\color{blue}↓})
& 26.0 ({\color{red}↑})
& 52.0 ({\color{blue}↓})
& 18.0 ({\color{red}↑})
& 0.0 ({\color{gray}---}) \\

Claude (BASE)
& 6.0 ({\color{gray}---})
& 2.0 ({\color{blue}↓})
& 92.0 ({\color{red}↑})
& 0.0 ({\color{gray}---})
& 0.0 ({\color{gray}---}) \\

Claude\_Empathy Prompt
& 10.0 ({\color{red}↑})
& 10.0 ({\color{gray}---})
& 80.0 ({\color{blue}↓})
& 0.0 ({\color{gray}---})
& 0.0 ({\color{gray}---}) \\

Claude\_Rephrase
& 0.0 ({\color{blue}↓})
& 8.0 ({\color{blue}↓})
& 92.0 ({\color{red}↑})
& 0.0 ({\color{gray}---})
& 0.0 ({\color{gray}---}) \\

\bottomrule
\end{tabular}}
\end{table*}
Taken together, the results suggest that affective misalignment is more pronounced in institutionally curated medical explanations (MedQuAD) than in conversational clinical exchanges (iCliniqQAs). While empathy prompting and rephrasing consistently reduce extreme negativity and increase neutrality across both datasets, certain architectures introduce an independent tendency toward positive reinforcement, revealing a systematic stylistic shift rather than strict replication of physician affective norms.

\subsubsection{Emotion Distribution}
\begin{figure}[t]
    \centering
    \includegraphics[width=\textwidth]{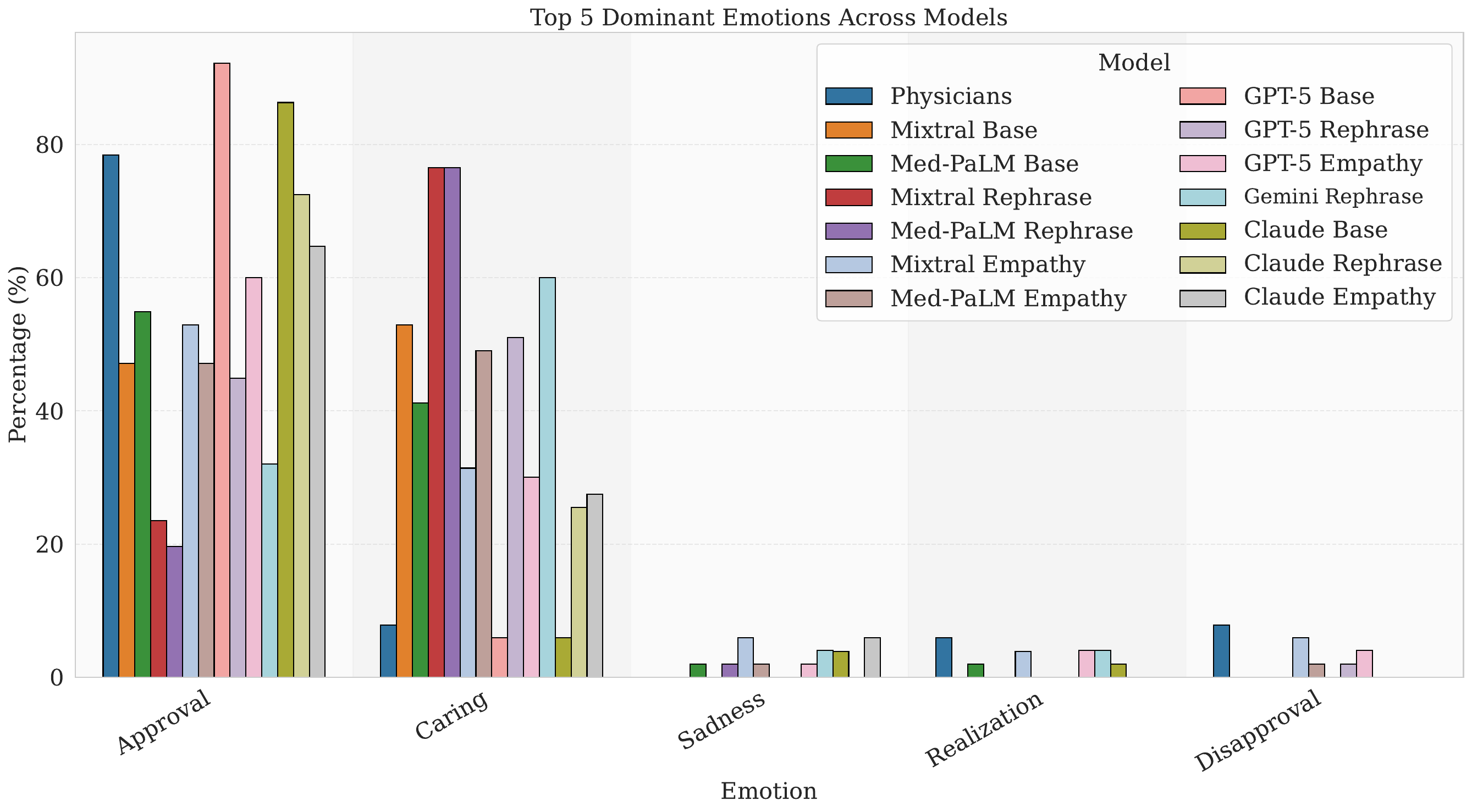}
    \caption{Emotions most frequently expressed by models on the MedQuAD dataset.}
    \label{fig:top5_emotions}
\end{figure}

\begin{figure}[t]
    \centering
    \includegraphics[width=\textwidth]{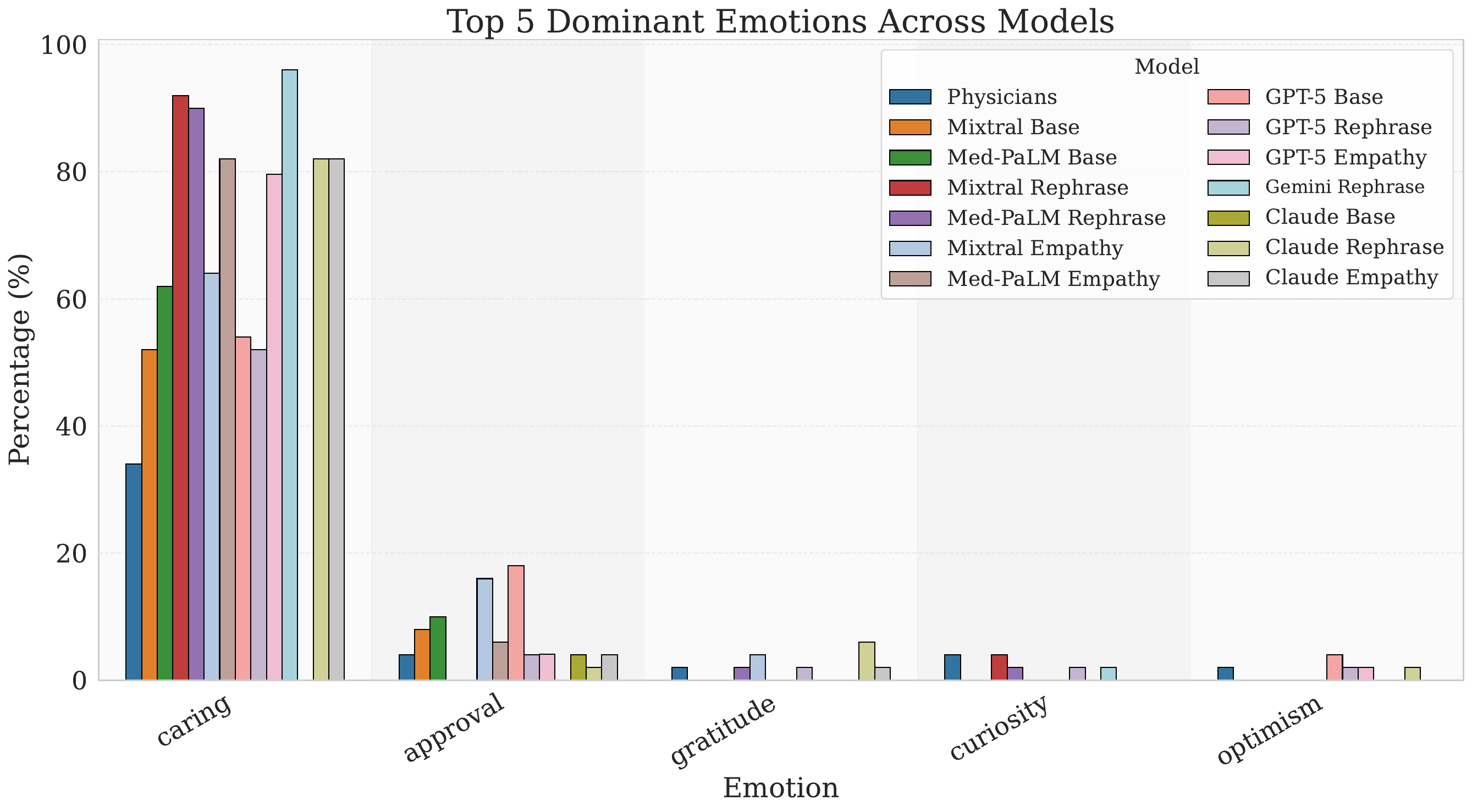}
    \caption{Emotions most frequently expressed by models on the iCliniqQAs dataset.}
    \label{fig:top5_emotions1}
\end{figure}

Beyond general sentiment, we analyzed the presence of 28 fine-grained emotional categories.
Figures~\ref{fig:top5_emotions} and~\ref{fig:top5_emotions1} report the five most frequent dominant emotions across systems in the MedQuAD and iCliniqQAs datasets, respectively.

Across both datasets, two emotions consistently dominate model-generated outputs: \textit{approval} and \textit{caring}. However, their relative balance differs substantially between datasets, reflecting the distinct communicative setting. In MedQuAD, physician-authored responses are strongly approval-oriented, with \textit{approval} as the dominant emotion in 78.4\% of cases, while \textit{caring} and \textit{disapproval} each account for 7.8\%, and \textit{realization} for 5.9\%. This pattern is consistent with institutional medical explanations, where clinicians primarily convey validation and guidance, with occasional corrective or reflective cues.

In contrast, iCliniqQAs exhibits a marked shift toward affective support. Here, physician answers are predominantly \textit{caring}-oriented (33.3\%), while \textit{approval} becomes secondary (2.0\%). The remaining dominant emotions appear at much lower rates, including \textit{gratitude} (2.0\%), \textit{curiosity} (3.9\%), and \textit{optimism} (2.0\%). This difference indicates that conversational consultations elicit a substantially more supportive and relational emotional style than standardized institutional explanations, even in physician-written content.

Several systematic model behaviors emerge across both datasets. First, LLMs display high variability in the expression of \textit{approval} on MedQuAD. Some base configurations are more approval-heavy than physicians, such as \textit{GPT5\_BASE} (92.23\%) and \textit{Claude\_BASE} (86.31\%), whereas others substantially reduce approval when prompted for caring or rewriting: \textit{Mixtral\_Rephrase} (23.5\%) and \textit{MedPaLM\_Rephrase} (19.60\%) illustrate a strong reallocation away from validation toward more explicitly supportive framing. In iCliniqQAs, approval is instead generally attenuated across systems and rarely becomes dominant; when it appears among the top emotions, it does so at modest levels (e.g., \textit{Mixtral\_Empathy} 16.00\%, \textit{GPT5\_Rephrase} 18.00\%), consistent with the dataset’s baseline emphasis on reassurance rather than endorsement.

Second, \textit{caring} is systematically amplified in LLM outputs relative to physicians in both datasets, but the magnitude of amplification depends on the conversational context. In MedQuAD, caring is dominant in only 7.8\% of physician responses, yet it becomes one of the primary emotions in most model settings, especially under empathy prompting and rewriting: \textit{Mixtral\_Empathy} (52.90\%) and \textit{MedPaLM\_Empathy} (41.20\%) strongly exceed physicians, while rewriting further accentuates caring, with \textit{Mixtral\_Rephrase} and \textit{MedPaLM\_Rephrase} reaching 76.5\%. In iCliniqQAs, the same tendency persists but starts from a substantially higher human baseline (33.30\%). Many models push caring to near-saturation levels, particularly in base configurations (e.g., \textit{Mixtral\_Rephrase} =  92.00\%, \textit{Claude\_Base} = 82.00\%, \textit{GPT5\_Base} = 92.0\%), indicating that in naturally emotional patient narratives, models converge toward a highly supportive stance regardless of whether they are explicitly prompted for empathy.

Third, negative or corrective emotions are attenuated in model outputs, especially in MedQuAD. In the first dataset, \textit{disapproval} is consistently present in physician texts (7.8\%) yet appears marginally or disappears in most LLM configurations, rarely exceeding 5.9\% and often remaining absent in the top emotions. This aligns with an avoidance of negatively directive stances in machine-generated clinical communication. In iCliniqQAs, disapproval is not among the dominant emotions for physicians or models; instead, low-frequency positive-affiliative emotions such as \textit{gratitude}, \textit{optimism}, and \textit{curiosity} emerge among the top categories, but remain limited in prevalence (generally below $\sim$5\%), suggesting that the overall affective profile is still largely governed by caring.

Taken together, fine-grained emotion analysis shows that LLMs do not reproduce physician affective behavior verbatim. Rather, they exhibit a systematic reweighting of emotional cues that amplifies affiliative signals such as \textit{caring} and, depending on the dataset, either preserves or reduces \textit{approval}. Importantly, the direction of this shift is dataset-dependent: institutional explanations (MedQuAD) highlight a transition from approval-dominant physician discourse toward caring-heavy model outputs, whereas real-world consultations (iCliniqQAs) already start from a caring-oriented physician baseline and are further pushed by LLMs toward near-uniform supportive affect. This difference reflects variation in emotional style and emphasis across contexts, rather than an absolute improvement in communication quality.

\begin{takeaway}
$R_{\text{LLM}}$ can approximate $R_{\text{Phys}}$ in overall emotional restraint 
across both datasets. However, fine-grained emotion analysis reveals a systematic 
reweighting rather than faithful replication: LLMs consistently amplify affiliative 
signals such as \textit{caring}, while attenuating corrective or discordant cues. 
This shift is dataset-dependent: models move from approval-dominant discourse in 
institutional texts to near-saturated caring in conversational settings, indicating 
stylistic modulation rather than improved clinical alignment.
\end{takeaway}

\subsection{RQ2 – Readability Analysis}
\begin{figure*}[t]
    \centering
    
    \begin{subfigure}[t]{\textwidth}
        \centering
        \includegraphics[width=\textwidth]{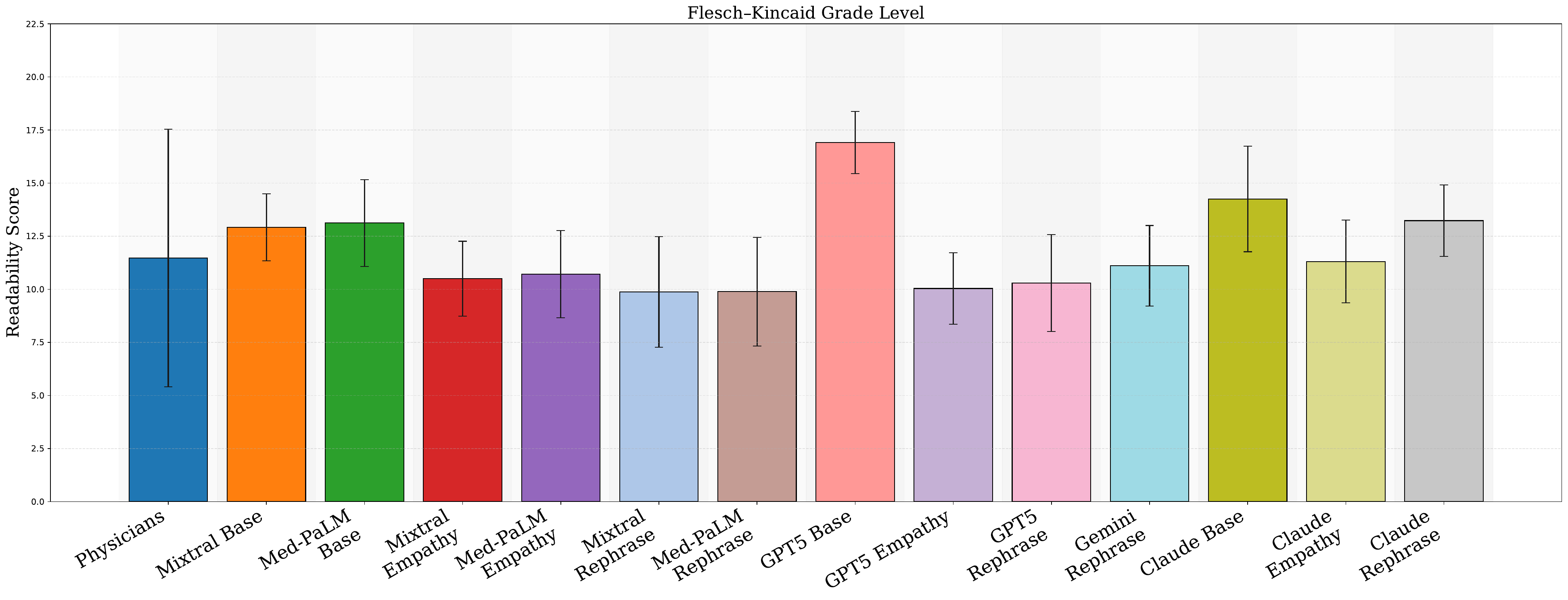}
        \caption{FKGL. Lower values = better readability.}
        \label{fig:fkgl}
    \end{subfigure}
    
    \vspace{1.5em} 
    
    \begin{subfigure}[t]{\textwidth}
        \centering
        \includegraphics[width=\textwidth]{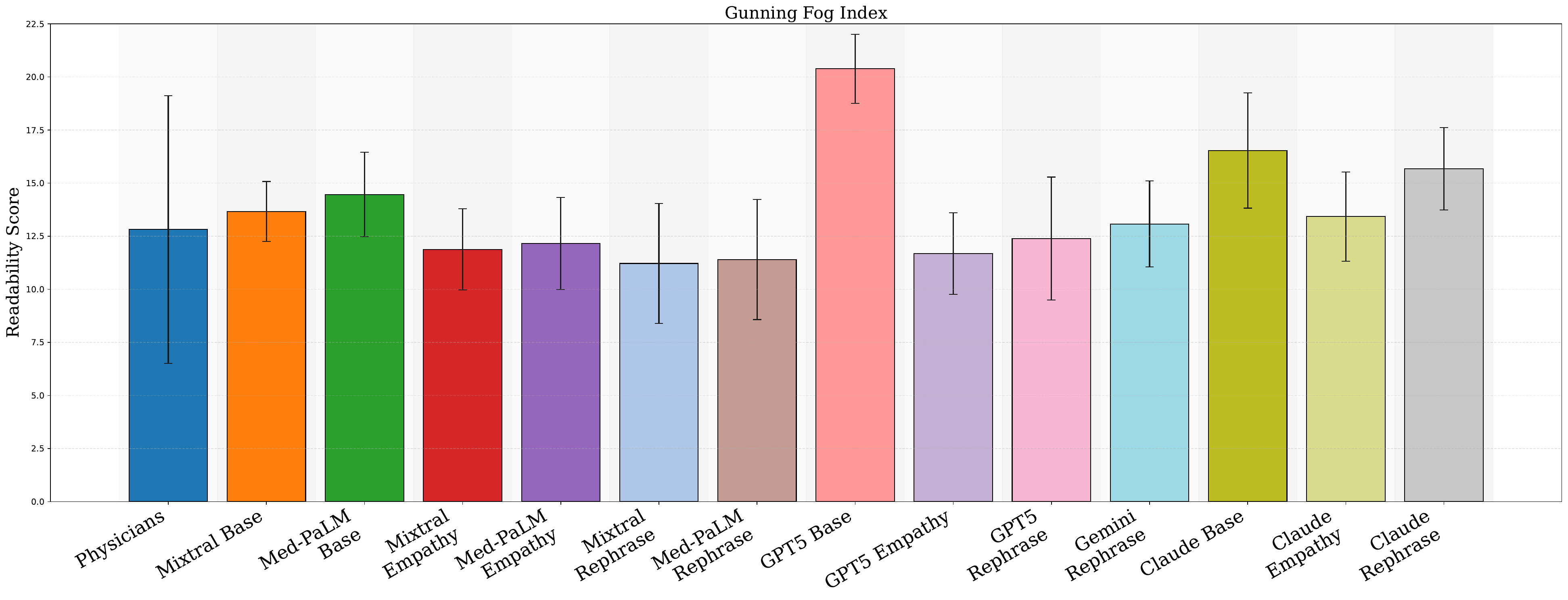}
        \caption{GFI. Higher values = harder text.}
        \label{fig:gfi}
    \end{subfigure}
    
    \caption{Readability analysis: Flesch–Kincaid Grade Level and Gunning Fog Index scores across physician and LLM outputs on the MedQuAD dataset.}
    \label{fig:readability}
\end{figure*}

\begin{figure*}[t]
    \centering
    
    \begin{subfigure}[t]{\textwidth}
        \centering
        \includegraphics[width=\textwidth]{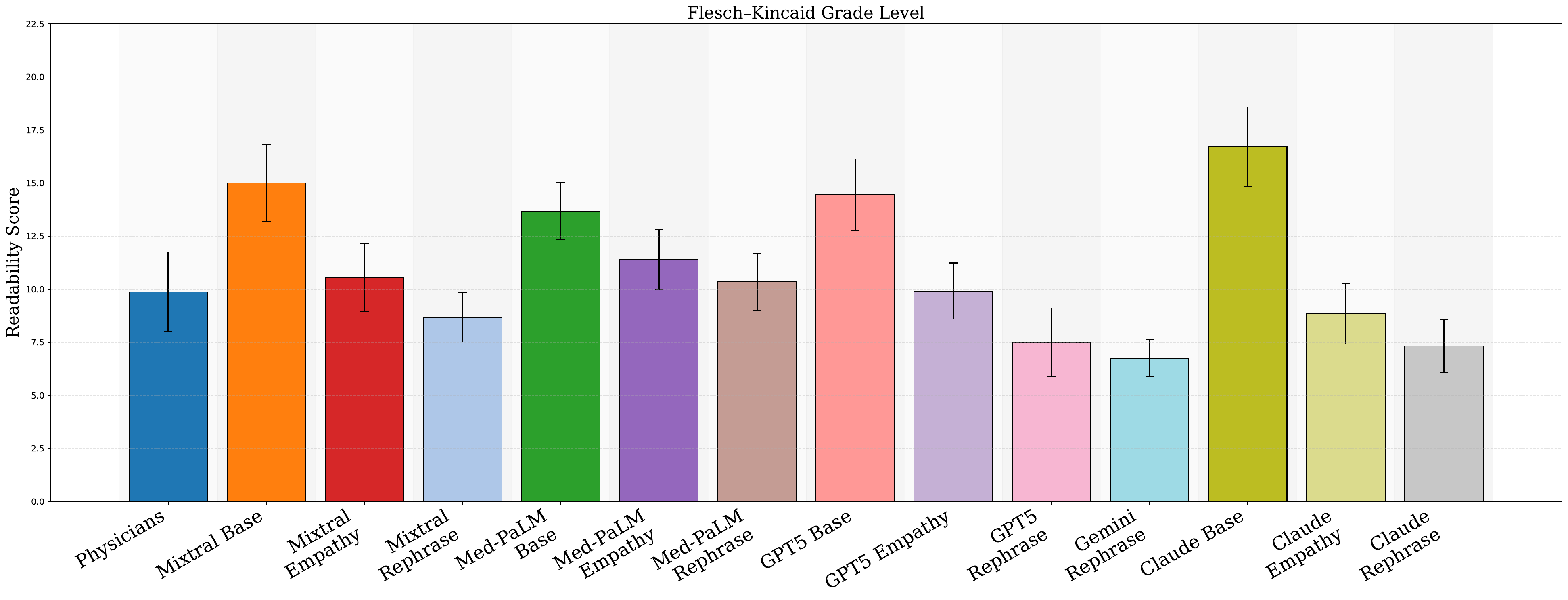}
        \caption{FKGL. Lower values = better readability.}
        \label{fig:fkgl1}
    \end{subfigure}
    
    \vspace{1.5em} 
    
    \begin{subfigure}[t]{\textwidth}
        \centering
        \includegraphics[width=\textwidth]{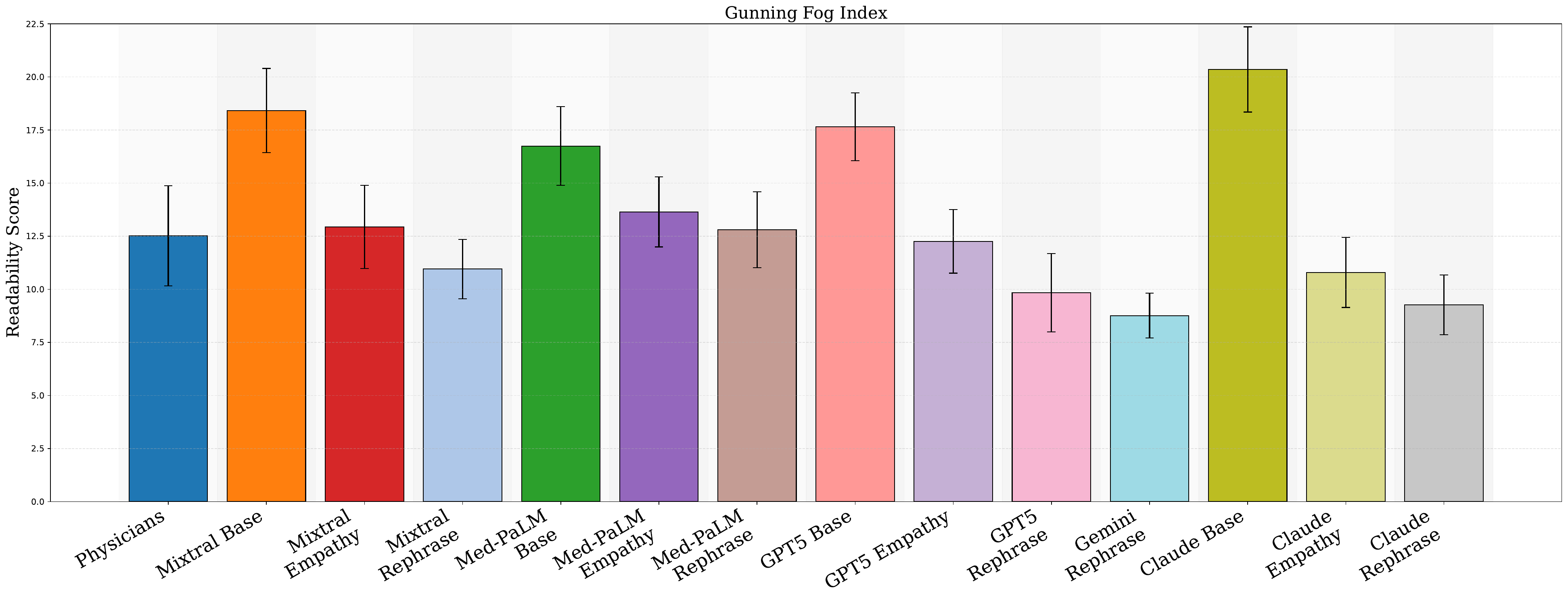}
        \caption{GFI. Higher values = harder text.}
        \label{fig:gfi1}
    \end{subfigure}
    
    \caption{Readability analysis: Flesch–Kincaid Grade Level and Gunning Fog Index scores across physician and LLM outputs on the iCliniqQAs dataset.}
    \label{fig:readability1}
\end{figure*}

This research question explores whether LLMs can produce more readable responses than those authored by physicians, who may rely on complex phrasing and technical jargon.
To evaluate the readability of each response type, we applied the FKGL and GFI metrics previously introduced in the \textit{Background} section. Figures~\ref{fig:readability} and~\ref{fig:readability1} report average scores for physician-written content and base (i.e., non–prompt-engineered, non–rephrased) LLM generations.

\begin{hyp}
\textit{Let $R_{\text{LLM}}^{\text{base}}$ be a zero-shot (baseline) LLM response without domain prompting or rewriting, and let $Read(\cdot)$ be a function that measures the readability of a text where lower scores indicate greater accessibility. We hypothesize that baseline LLM generations will exhibit \textbf{equal or higher readability} compared to physician-authored content $R_{\text{Phys}}$, i.e.,}
\[
\mathbb{E}[Read(R_{\text{LLM}}^{\text{base}})] \leq \mathbb{E}[Read(R_{\text{Phys}})].
\]
\end{hyp}

Overall, physician-authored responses exhibit moderate complexity in both datasets.
On MedQuAD, physicians show FKGL $= 11.47$ and GFI $= 12.82$.
On iCliniqQAs, physicians exhibit FKGL $= 12.50$ and GFI $= 12.60$, indicating that conversational physician responses are not substantially simpler than institutional ones in terms of formal grade-level metrics.

\begin{figure}[t]
    \centering
    \includegraphics[width=\linewidth]{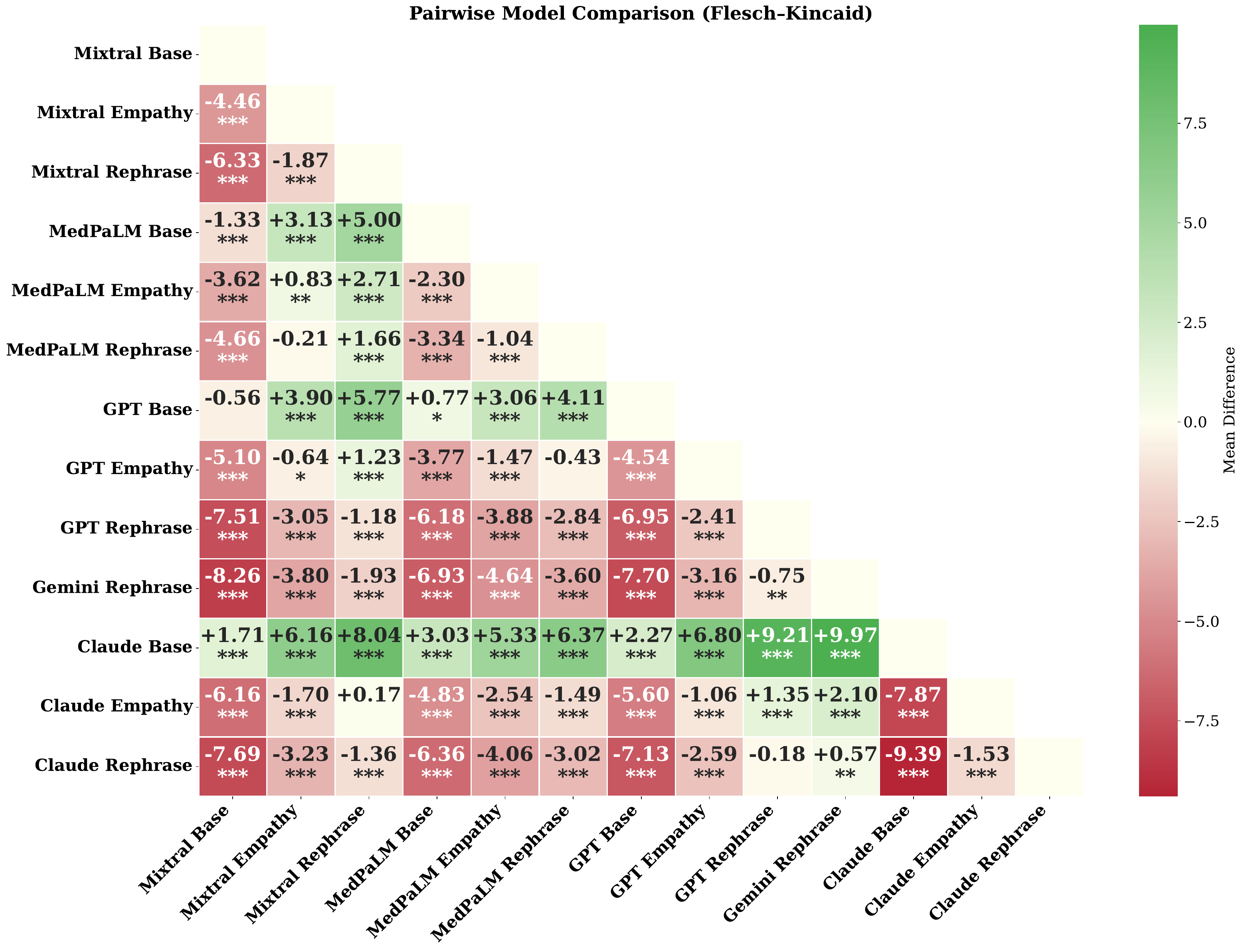}
    \caption{
    Pairwise differences in Flesch–Kincaid Grade Level (FKGL) across systems on the  iCliniqQAs dataset.
    Each cell reports the mean difference between row and column models (row minus column);
    negative values indicate better readability for the row model.
    Statistical significance is assessed via paired $t$-tests with Benjamini--Hochberg FDR correction
    (\textit{*} $p<0.05$, \textit{**} $p<0.01$, \textit{***} $p<0.001$).}
    \label{fig:readability_fk_heatmap1}
\end{figure}

\begin{figure}[t]
    \centering
    \includegraphics[width=\linewidth]{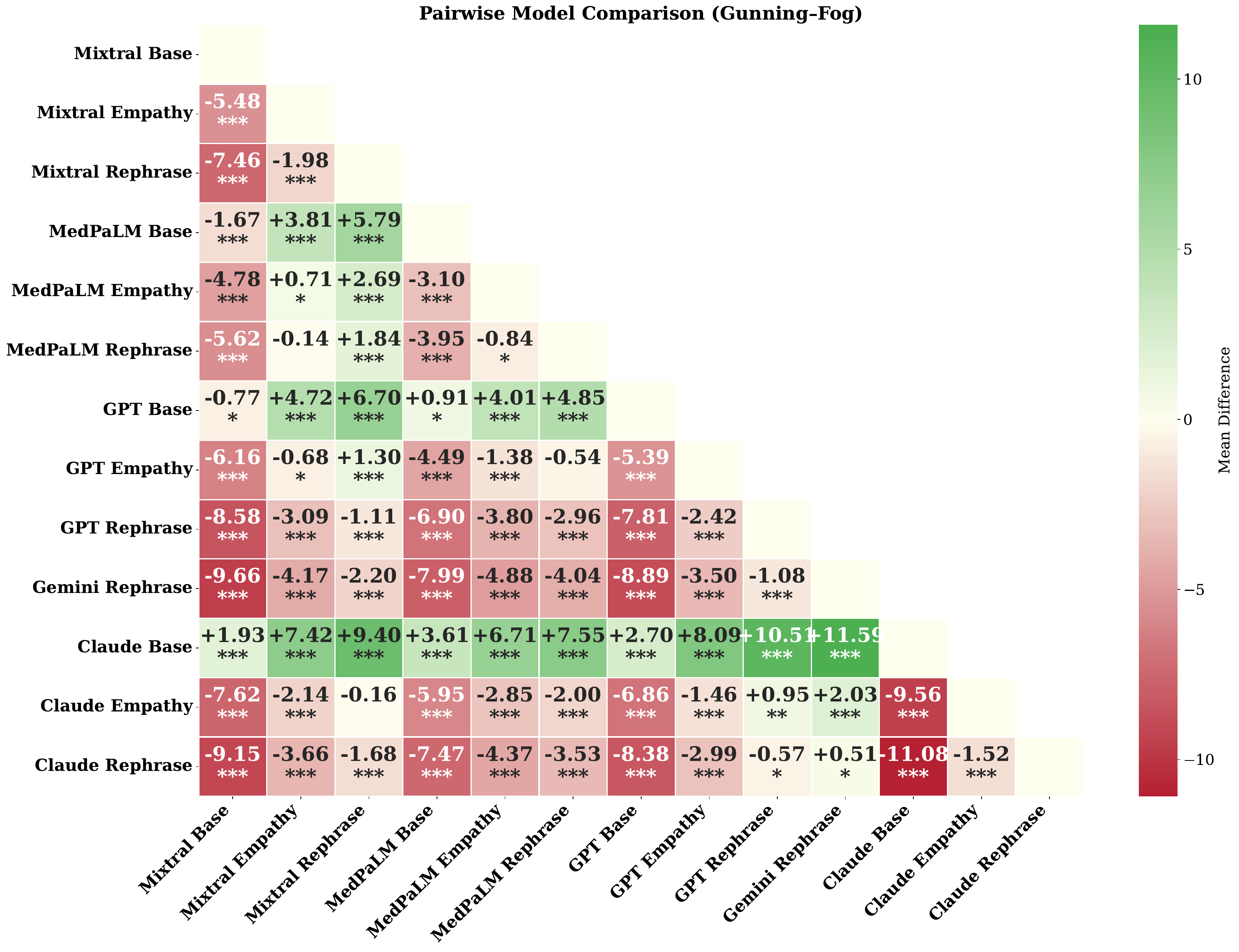}
    \caption{
    Pairwise differences in Gunning Fog Index (GFI) across systems on the  iCliniqQAs dataset.
    Values represent mean score differences (row minus column); lower values correspond to
    easier-to-read text.
    Significance is evaluated using paired $t$-tests with FDR correction
    (\textit{*} $p<0.05$, \textit{**} $p<0.01$, \textit{***} $p<0.001$).
    }
    \label{fig:readability_gf_heatmap1}
\end{figure}

In MedQuAD, \textit{GPT5\_Base} displays substantially higher complexity 
(FKGL $= 16.91$, GFI $= 20.39$), and \textit{Claude\_Base} also exceeds 
physician readability (FKGL $= 14.26$, GFI $= 16.54$). 
\textit{Mixtral\_Base} and \textit{MedPaLM\_Base} remain closer to physician levels.
For the iCliniqQAs dataset, \textit{GPT5\_Base} produces highly complex text 
(FKGL $= 17.60$, GFI $= 17.60$), while \textit{Claude\_Base} reaches the highest 
GFI values overall (GFI $= 20.30$). 
As shown in Figures~\ref{fig:readability_fk_heatmap1} 
and~\ref{fig:readability_gf_heatmap1}, on iCliniqQAs \textit{GPT5\_Base} is 
significantly less readable than physicians across both metrics 
($\Delta$FKGL $= +5.44$, $\Delta$GFI $= +7.57$, all $p < 0.001$, FDR-corrected).
\textit{Claude\_Base} also produces significantly more complex text than physicians 
($\Delta$FKGL $= +2.78$, $\Delta$GFI $= +3.71$, $p < 0.01$).
In contrast, \textit{Mixtral\_Base} and \textit{MedPaLM\_Base} do not show 
statistically significant deviations from physician readability on either dataset, 
confirming that their baseline lexical complexity is broadly aligned with 
expert-authored responses.

\begin{figure}[t]
    \centering
    \includegraphics[width=\linewidth]{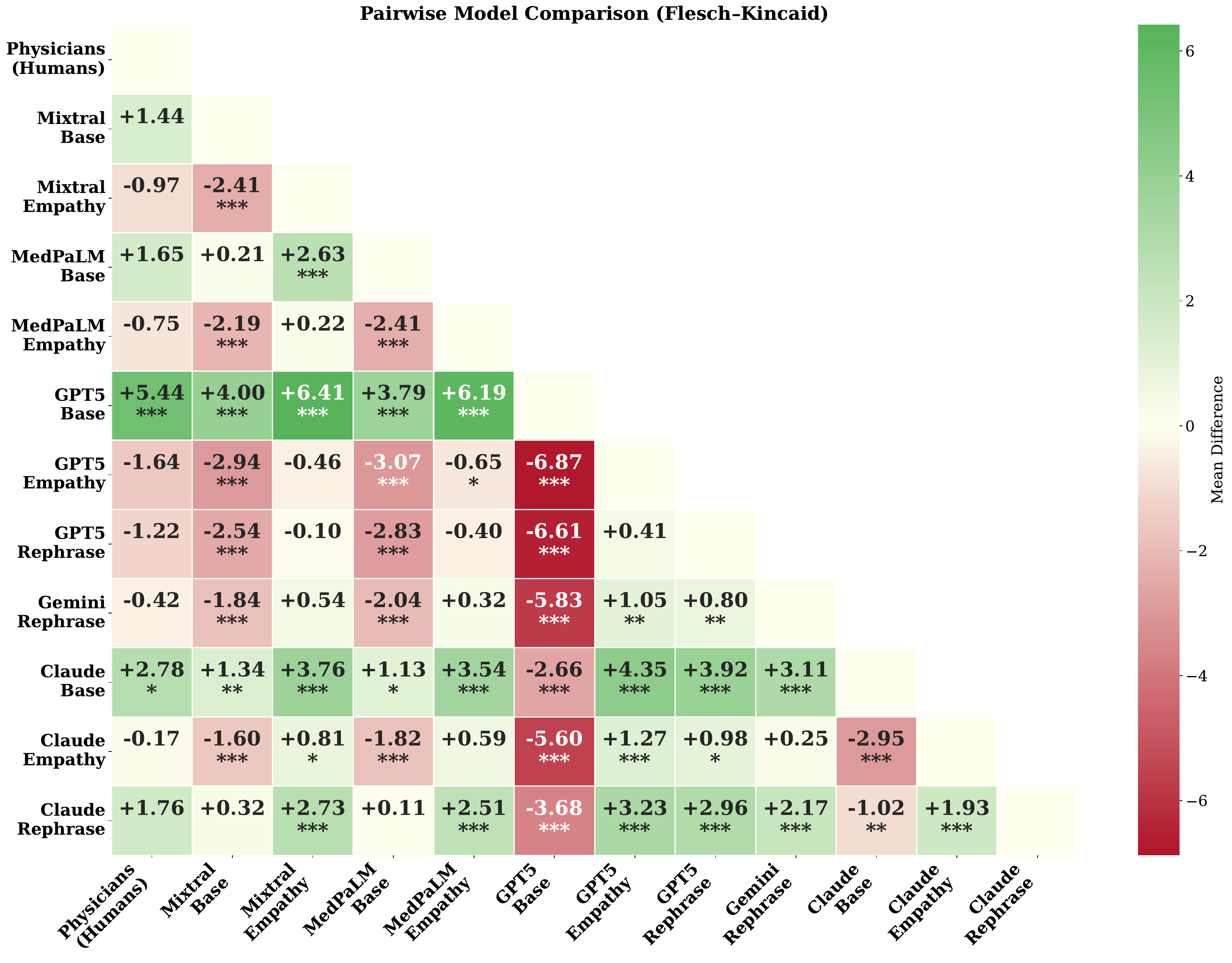}
    \caption{
    Pairwise differences in Flesch–Kincaid Grade Level (FKGL) across systems on the MedQuAD dataset.
    Each cell reports the mean difference between row and column models (row minus column);
    negative values indicate better readability for the row model.
    Statistical significance is assessed via paired $t$-tests with Benjamini--Hochberg FDR correction
    (\textit{*} $p<0.05$, \textit{**} $p<0.01$, \textit{***} $p<0.001$).}
    \label{fig:readability_fk_heatmap}
\end{figure}

\begin{figure}[t]
    \centering
    \includegraphics[width=\linewidth]{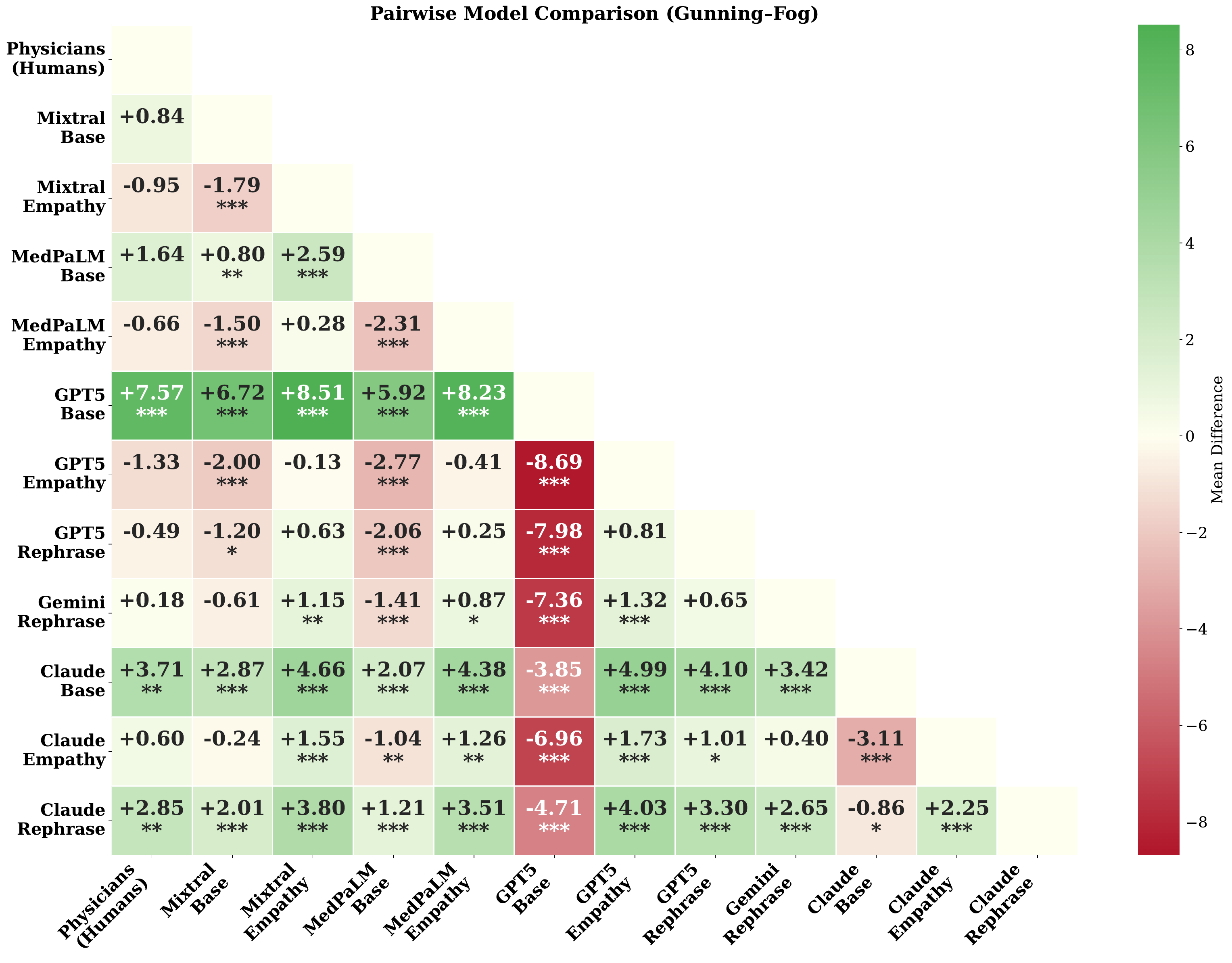}
    \caption{
    Pairwise differences in Gunning Fog Index (GFI) across systems on the MedQuAD dataset.
    Values represent mean score differences (row minus column); lower values correspond to easier-to-read text.
    Significance is evaluated using paired $t$-tests with FDR correction
    (\textit{*} $p<0.05$, \textit{**} $p<0.01$, \textit{***} $p<0.001$).
    }
    \label{fig:readability_gf_heatmap}
\end{figure}

Across both datasets, empathy prompting and rephrasing systematically reduce 
linguistic complexity relative to baseline models.
On MedQuAD (Figures~\ref{fig:readability_fk_heatmap} 
and~\ref{fig:readability_gf_heatmap}), \textit{Mixtral\_Empathy} and 
\textit{MedPaLM\_Empathy} reduce FKGL by $-2.41$ and $-2.41$ points respectively 
compared to their base variants, with analogous improvements in GFI 
($-1.79$ and $-2.31$, all $p < 0.001$).
For \textit{GPT5}, the reduction is even larger: \textit{GPT5\_Empathy} and 
\textit{GPT5\_Rephrase} lower FKGL by $-6.87$ and $-6.61$ points and GFI by 
$-8.69$ and $-7.98$ points relative to \textit{GPT5\_Base} (all $p < 0.001$).
\textit{Claude\_Empathy} and \textit{Claude\_Rephrase} also improve readability 
relative to \textit{Claude\_Base} ($\Delta$FKGL $= -2.95$ and $-1.02$; 
$\Delta$GFI $= -3.11$ and $-0.86$, $p < 0.01$).

A comparable pattern is observed in iCliniqQAs
(Figures~\ref{fig:readability_fk_heatmap1} and~\ref{fig:readability_gf_heatmap1}).
\textit{GPT5\_Empathy} reduces FKGL by $-4.54$ and GFI by $-5.39$ relative to 
\textit{GPT5\_Base} (both $p < 0.001$), and \textit{GPT5\_Rephrase} yields even 
larger improvements ($\Delta$FKGL $= -6.95$, $\Delta$GFI $= -7.81$, $p < 0.001$).
The effect is especially pronounced for \textit{Claude}: \textit{Claude\_Empathy} 
lowers FKGL by $-7.87$ and GFI by $-9.56$ relative to \textit{Claude\_Base}, while 
\textit{Claude\_Rephrase} further reduces complexity ($\Delta$FKGL $= -9.39$, 
$\Delta$GFI $= -11.08$, all $p < 0.001$).

\textit{Gemini\_Rephrase} also shows statistically significant improvements 
relative to more complex baselines (e.g., $\Delta$FKGL $= -3.16$, 
$\Delta$GFI $= -3.50$ vs.\ \textit{GPT5\_Base} on iCliniqQAs, $p < 0.001$).

Taken together, these findings confirm that improved readability is not an intrinsic property of LLM output. 
Baseline generations from \textit{Mixtral} and \textit{MedPaLM} approximate physician readability across both datasets, whereas \textit{GPT5\_Base} and \textit{Claude\_Base} produce significantly more complex prose.
Consistent readability gains emerge primarily when models are explicitly instructed or used as rewriting assistants, indicating that accessibility depends strongly on prompting strategy rather than architecture alone.

\begin{takeaway}
Baseline LLM generations do not systematically improve accessibility across datasets. 
While \textit{Mixtral} and \textit{Med-PaLM} approximate physician readability in both MedQuAD and iCliniqQAs, \textit{GPT5\_Base} and \textit{Claude\_Base} consistently produce significantly more complex text than clinicians. 
Substantial and statistically significant readability gains emerge only under empathy prompting or collaborative rewriting. 
Accessibility therefore appears to be a controllable property induced by alignment strategies rather than an intrinsic characteristic of large language models.
\end{takeaway}

\subsection{RQ3 – Effect of Prompt Engineering on AI Alignment}
This research question assesses whether empathy-enhancing prompt design can steer LLMs toward more emotionally appropriate and readable outputs.

\begin{hyp}
\textit{
Let $R_{\text{LLM\_Empathy Prompt}}$ denote responses generated under the empathy-enhanced Empathy Prompt condition.
We hypothesize that Empathy Prompt primarily affects affective and communicative style rather than technical content,
leading to (i) increased affective support and (ii) improved readability due to indirect stylistic
simplification rather than explicit textual optimization, relative to zero-shot outputs, i.e.,
\[
\begin{aligned}
(i)\quad &\mathbb{E}[E(R_{\text{LLM\_Empathy Prompt}})] 
      > \mathbb{E}[E(R_{\text{LLM\_Base}})]
      \quad\text{and}\\[4pt]
(ii)\quad &\mathbb{E}[Read(R_{\text{LLM\_Empathy Prompt}})] 
      < \mathbb{E}[Read(R_{\text{LLM\_Base}})] .
\end{aligned}
\]
}
\end{hyp}

Readability outcomes are reported in Figures~\ref{fig:readability} and~\ref{fig:readability1}, which present average Flesch–Kincaid Grade Level (FKGL) and Gunning Fog Index (GFI) scores for physician-authored responses and for each model configuration across both datasets.

Across models, Empathy Prompt consistently lowers FKGL and GFI scores relative to their corresponding base variants in MedQuAD. 
For instance, \textit{Mixtral\_Empathy Prompt} reduces FKGL from $12.91$ to $11.88$ and GFI from $13.66$ to $12.50$, while \textit{MedPaLM\_Empathy Prompt} shows similar reductions (FKGL $13.13 \rightarrow 10.72$; GFI $14.47 \rightarrow 12.16$). 
For larger architectures, the effect is even more pronounced: \textit{GPT5\_Empathy} lowers FKGL from $16.91$ to $10.04$ and GFI from $20.39$ to $11.70$, representing reductions of $-6.87$ and $-8.69$ points respectively (all $p < 0.001$). 
These results indicate that empathy-oriented prompting encourages simpler sentence construction and reduced lexical density in institutionally curated medical explanations.

A comparable but context-sensitive pattern emerges in iCliniqQAs. 
Here, physicians exhibit FKGL $= 12.50$ and GFI $= 12.60$. 
\textit{GPT5\_Base} produces substantially higher complexity (FKGL $= 17.60$, GFI $= 17.60$), whereas \textit{GPT5\_Empathy} reduces these scores to FKGL $= 13.06$ and GFI $= 12.21$, yielding reductions of $-4.54$ and $-5.39$ points respectively (both $p < 0.001$). 
\textit{Claude\_Empathy} similarly improves readability compared to \textit{Claude\_Base} ($\Delta$FKGL $= -7.87$, $\Delta$GFI $= -9.56$, $p < 0.001$).
Thus, across both institutional (MedQuAD) and conversational (iCliniqQAs) settings, empathy prompting systematically reduces linguistic complexity.

In terms of sentiment alignment (Tables~\ref{tab:sentiment_distribution_transposed} and~\ref{tab:sentiment_distribution_second_dataset}), Empathy Prompt shifts model responses toward more neutral and less confrontational phrasing in MedQuAD. 
\textit{Mixtral\_Empathy Prompt} increases Neutral responses from $56.86\%$ to $74.51\%$ while reducing Very Negative sentiment from $43.14\%$ to $23.53\%$. 
\textit{MedPaLM\_Empathy Prompt} shows a comparable shift (Very Negative $45.10\% \rightarrow 25.49\%$; Neutral $41.18\% \rightarrow 68.63\%$). 
\textit{GPT5\_Empathy Prompt} reduces Very Negative sentiment from $40.00\%$ to $22.00\%$ and increases Neutral responses from $58.00\%$ to $74.00\%$.

In iCliniqQAs, baseline physician sentiment is already strongly Neutral ($84.00\%$) with low Very Negative content ($6.00\%$). 
In this setting, empathy prompting reduces extreme negativity but does not universally increase neutrality. 
For example, \textit{Mixtral\_Empathy Prompt} reduces Very Negative responses from $14.00\%$ to $2.00\%$, while maintaining Neutral at $84.00\%$. 
\textit{GPT5\_Empathy Prompt}, however, reduces Very Negative from $28.00\%$ to $16.00\%$ while shifting distribution toward both Negative ($16.00\%$) and Neutral ($62.00\%$), indicating that affective modulation in conversational data is more architecture-dependent than in MedQuAD. 
\textit{Claude\_Empathy Prompt} slightly increases Very Negative responses from $6.00\%$ to $10.00\%$, demonstrating that empathy prompting does not uniformly guarantee improved sentiment alignment in patient-facing dialogue.

Empathy Prompt increases supportive emotional cues without artificially inflating Positive sentiment in MedQuAD, where Positive remains at $0.00\%$ for most systems. 
In contrast, in iCliniqQAs, certain configurations introduce modest Positive proportions (e.g., \textit{Mixtral\_Empathy Prompt} $= 4.00\%$, \textit{GPT5\_Empathy Prompt} $= 4.00\%$), reflecting the conversational tone of the dataset.

Fine-grained emotion analysis (Figures~\ref{fig:top5_emotions} and~\ref{fig:top5_emotions1}) further clarifies this divergence. 
In MedQuAD, empathy prompting substantially amplifies \textit{caring} relative to physicians (from $7.80\%$ to over $40.00\%$ in several configurations), whereas in iCliniqQAs, where physician responses are already caring-dominant ($33.30\%$), models often push caring toward near-saturation levels (e.g., \textit{Mixtral\_Rephrase} $= 92.00\%$). 
Thus, the emotional effect of empathy prompting is additive in institutional discourse but saturating in conversational settings.

Statistical testing via paired $t$-tests with Benjamini–Hochberg FDR correction confirms that Empathy Prompt produces significant improvements over baseline generations in readability across both datasets (all $p < 0.01$ for major architectures). 
However, sentiment improvements are dataset-dependent: reductions in extreme negativity are systematic in MedQuAD but more variable in iCliniqQAs, where baseline physician affect is already strongly neutral and supportive.

Taken together, these results indicate that empathy prompting robustly enhances readability in both institutional and conversational medical communication. 
Its effect on affective alignment, however, is moderated by the underlying discourse context: it corrects polarity amplification in formal explanatory texts but produces more architecture-specific shifts in already supportive patient–physician dialogue.

\begin{takeaway}
Empathy-enhanced prompting (Empathy Prompt) systematically improves readability and modulates affective tone across both datasets. 
It reduces extreme negativity and increases affiliative cues --- particularly \textit{caring} --- while lowering linguistic complexity relative to baseline generations. 
Readability gains are especially pronounced for larger architectures such as \textit{GPT5}, whereas sentiment shifts are more dataset-dependent: polarity correction is consistent in institutional texts (MedQuAD) but more variable in conversational dialogue (iCliniqQAs). 
Overall, Empathy Prompt acts as a controllable alignment mechanism, steering emotional tone toward clinical norms while improving accessibility without compromising semantic fidelity.
\end{takeaway}

\subsection{RQ4 – Human–AI Collaboration}
This research question evaluates LLMs not only as content generators, but also as editors capable of revising expert-authored medical texts to improve clarity and emotional resonance.

\begin{hyp}
\textit{
Let $R_{\text{LLM\_Rephrase}}$ denote the physician-authored response rewritten by an 
LLM using the Rephrase prompt, and $R_{\text{Phys}}$ the original physician response. 
We hypothesize that collaborative rewriting will produce responses that are (i) \textbf{more readable} and (ii) \textbf{more affectively supportive} than the original 
physician-authored text, i.e.,
\[
\begin{aligned}
(i)\quad &\mathbb{E}[Read(R_{\text{LLM\_Rephrase}})] 
      < \mathbb{E}[Read(R_{\text{Phys}})]
      \quad\text{and}\\[4pt]
(ii)\quad &\mathbb{E}[E(R_{\text{LLM\_Rephrase}})] 
      > \mathbb{E}[E(R_{\text{Phys}})].
\end{aligned}
\]
}
\end{hyp}

Rewriting systematically shifts emotional polarity toward more supportive and less confrontational phrasing across both datasets.

In MedQuAD (Table~\ref{tab:sentiment_distribution_transposed}), rephrase variants markedly reduce \textit{Very Negative} sentiment while increasing \textit{Neutral} responses. 
\textit{MedPaLM\_Rephrase} achieves the strongest moderation effect (\textit{Very Negative} = 19.61\%, \textit{Neutral} = 72.55\%), followed closely by \textit{Mixtral\_Rephrase} (\textit{Very Negative} = 21.57\%, \textit{Neutral} = 70.59\%). 
\textit{GPT5\_Rephrase} reduces \textit{Very Negative} sentiment from 40.00\% to 22.00\% while increasing neutrality to 68.00\%. 
Although \textit{Claude\_Rephrase} remains more polarity-heavy than other models (\textit{Very Negative} = 56.00\%), it still substantially moderates its baseline behavior relative to \textit{Claude\_Base} (82.00\%).

A different but structurally consistent pattern emerges in iCliniqQAs (Table~\ref{tab:sentiment_distribution_second_dataset}). 
Here, physician responses already exhibit strong neutrality (\textit{Neutral} = 84.00\%, \textit{Very Negative} = 6.00\%), reflecting the conversational nature of the dataset. 
In this context, rewriting primarily compresses extreme negativity and increases neutral dominance rather than correcting polarity amplification. 
\textit{Mixtral\_Rephrase} achieves \textit{Very Negative} = 0.00\% and \textit{Neutral} = 90.00\%, while \textit{MedPaLM\_Rephrase} yields \textit{Very Negative} = 2.00\% and \textit{Neutral} = 90.00\%. 
\textit{GPT5\_Rephrase} reduces \textit{Very Negative} from 28.00\% to 0.00\% and increases \textit{Neutral} to 88.00\%. 
\textit{Claude\_Rephrase} similarly eliminates extreme negativity (\textit{Very Negative} = 0.00\%) and raises \textit{Neutral} to 92.00\%.

Emotional tone analysis (Figures~\ref{fig:top5_emotions} and~\ref{fig:top5_emotions1}) further supports these trends. 
In MedQuAD, rewriting substantially increases \textit{caring} relative to physicians (7.80\%), with \textit{Mixtral\_Rephrase} and \textit{MedPaLM\_Rephrase} reaching 76.50\%. 
In iCliniqQAs, where physician responses are already caring-dominant (33.30\%), rewriting pushes affective support toward near-saturation levels (e.g., \textit{Mixtral\_Rephrase} = 92.00\%). 
Thus, rewriting acts as polarity correction in institutional discourse and as affective amplification in conversational dialogue.

Rewriting also enhances linguistic accessibility in both datasets. 
On MedQuAD, \textit{Mixtral\_Rephrase} and \textit{MedPaLM\_Rephrase} reduce FKGL from 12.91 to 11.18 and from 13.13 to 10.56, respectively, and GFI from 13.66 to 12.45 and from 14.47 to 12.16. 
\textit{GPT5\_Rephrase} lowers FKGL from 16.91 to 10.30 and GFI from 20.39 to 12.41, yielding statistically significant improvements (all $p < 0.001$). 
In iCliniqQAs, similar reductions are observed: \textit{GPT5\_Rephrase} decreases FKGL by $-6.95$ and GFI by $-7.81$ relative to \textit{GPT5\_Base}, while \textit{Claude\_Rephrase} reduces FKGL by $-9.39$ and GFI by $-11.08$ relative to \textit{Claude\_Base} (all statistically significant under FDR correction).

Notably, the magnitude of readability improvement is comparable across datasets, but the emotional effect differs in function: 
in MedQuAD, rewriting primarily mitigates excessive polarity, whereas in iCliniqQAs it consolidates an already supportive conversational tone.

Paired statistical tests with Benjamini–Hochberg False Discovery Rate correction confirm that rewriting introduces statistically significant improvements over baseline generations in sentiment distribution, fine-grained emotion profiles, and readability across multiple model families (all $p < 0.01$).

\begin{takeaway}
Collaborative rewriting consistently improves emotional alignment and linguistic accessibility across both datasets. 
It reduces extreme negativity, increases neutral and supportive phrasing, and lowers readability scores relative to baseline generations. 
Open-source models such as \textit{MedPaLM\_Rephrase} and \textit{Mixtral\_Rephrase} exhibit the most stable cross-dataset gains, while larger architectures (e.g., \textit{GPT5\_Rephrase}, \textit{Claude\_Rephrase}) demonstrate substantial reductions in polarity and complexity compared to their base variants. 
Rewriting therefore emerges as a robust post-hoc alignment mechanism that enhances clarity and affective appropriateness without degrading semantic fidelity.
\end{takeaway}

\begin{table*}[!htbp]
\centering
\caption{Mean ($\bar{x}$) and standard deviation ($\sigma$) of physician and patient (human) ratings on the MedQuAD. Arrows indicate comparison to Doctor: \textcolor{red}{$\uparrow$} = higher, \textcolor{blue}{$\downarrow$} = lower, \textcolor{gray}{-} = similar. Bold values denote best-performing variants per metric (excluding Physician baseline).}
\label{tab:rq5_dual_extended}
\resizebox{\textwidth}{!}{
\begin{tabular}{l|ccc|ccc}
\toprule
\multirow{2}{*}{\textbf{Response Variant}} &
\multicolumn{3}{c|}{\textbf{Expert Evaluation (Human)}} &
\multicolumn{3}{c}{\textbf{Patient Evaluation (Human)}} \\
& $\bar{x}_{\text{Accuracy}} \pm \sigma$ & $\bar{x}_{\text{Style}} \pm \sigma$ & $\bar{x}_{\text{Precision}} \pm \sigma$ 
& $\bar{x}_{\text{Trust}} \pm \sigma$ & $\bar{x}_{\text{Compr.}} \pm \sigma$ & $\bar{x}_{\text{Emot. Tone}} \pm \sigma$ \\
\midrule

\textbf{\textit{Physician Answer}} 
& $5.00 \pm 0.00$ & $5.00 \pm 0.00$ & $5.00 \pm 0.00$ 
& $2.50 \pm 0.50$ & $2.10 \pm 0.60$ & $2.30 \pm 0.50$ \\
\midrule

Mixtral                  
& $\mathbf{5.00 \pm 0.00}\,(\textcolor{gray}{-})$ 
& $4.50 \pm 0.00\,(\textcolor{blue}{\downarrow})$ 
& $\mathbf{5.00 \pm 0.00}\,(\textcolor{gray}{-})$
& $3.80 \pm 0.30\,(\textcolor{red}{\uparrow})$ 
& $4.10 \pm 0.20\,(\textcolor{red}{\uparrow})$ 
& $3.70 \pm 0.30\,(\textcolor{red}{\uparrow})$ \\

Med-PaLM                
& $\mathbf{5.00 \pm 0.00}\,(\textcolor{gray}{-})$ 
& $4.50 \pm 0.00\,(\textcolor{blue}{\downarrow})$ 
& $\mathbf{5.00 \pm 0.00}\,(\textcolor{gray}{-})$
& $3.50 \pm 0.30\,(\textcolor{red}{\uparrow})$ 
& $4.00 \pm 0.30\,(\textcolor{red}{\uparrow})$ 
& $3.60 \pm 0.20\,(\textcolor{red}{\uparrow})$ \\

Claude                  
& $\mathbf{5.00 \pm 0.00}\,(\textcolor{gray}{-})$ 
& $4.50 \pm 0.00\,(\textcolor{blue}{\downarrow})$ 
& $\mathbf{5.00 \pm 0.00}\,(\textcolor{gray}{-})$
& $3.40 \pm 0.20\,(\textcolor{red}{\uparrow})$ 
& $3.20 \pm 0.10\,(\textcolor{red}{\uparrow})$ 
& $3.10 \pm 0.10\,(\textcolor{red}{\uparrow})$ \\

GPT5                    
& $4.00 \pm 1.41\,(\textcolor{blue}{\downarrow})$ 
& $4.00 \pm 0.00\,(\textcolor{blue}{\downarrow})$ 
& $4.50 \pm 0.71\,(\textcolor{blue}{\downarrow})$
& $3.60 \pm 0.20\,(\textcolor{red}{\uparrow})$ 
& $3.20 \pm 0.20\,(\textcolor{red}{\uparrow})$ 
& $2.30 \pm 0.50\,(\textcolor{gray}{-})$ \\
\midrule

Mixtral\_Empathy\_Prompt         
& $4.00 \pm 1.41\,(\textcolor{blue}{\downarrow})$ 
& $4.50 \pm 0.71\,(\textcolor{blue}{\downarrow})$ 
& $4.00 \pm 1.41\,(\textcolor{blue}{\downarrow})$
& $4.60 \pm 0.20\,(\textcolor{red}{\uparrow})$ 
& $\mathbf{4.70 \pm 0.10\,(\textcolor{red}{\uparrow})}$ 
& $4.80 \pm 0.10\,(\textcolor{red}{\uparrow})$ \\

Med-PaLM\_Empathy\_Prompt        
& $4.50 \pm 0.71\,(\textcolor{blue}{\downarrow})$ 
& $4.50 \pm 0.71\,(\textcolor{blue}{\downarrow})$ 
& $4.50 \pm 0.71\,(\textcolor{blue}{\downarrow})$
& $4.40 \pm 0.20\,(\textcolor{red}{\uparrow})$ 
& $4.50 \pm 0.20\,(\textcolor{red}{\uparrow})$ 
& $4.50 \pm 0.20\,(\textcolor{red}{\uparrow})$ \\

Claude\_Empathy\_Prompt         
& $4.50 \pm 0.71\,(\textcolor{blue}{\downarrow})$ 
& $4.50 \pm 0.71\,(\textcolor{blue}{\downarrow})$ 
& $4.50 \pm 0.71\,(\textcolor{blue}{\downarrow})$
& $3.80 \pm 0.20\,(\textcolor{red}{\uparrow})$ 
& $3.60 \pm 0.20\,(\textcolor{red}{\uparrow})$ 
& $3.80 \pm 0.50\,(\textcolor{red}{\uparrow})$ \\

GPT5\_Empathy\_Prompt           
& $4.00 \pm 1.41\,(\textcolor{blue}{\downarrow})$ 
& $4.50 \pm 0.71\,(\textcolor{blue}{\downarrow})$ 
& $4.50 \pm 0.71\,(\textcolor{blue}{\downarrow})$
& $4.20 \pm 0.10\,(\textcolor{red}{\uparrow})$ 
& $3.50 \pm 0.30\,(\textcolor{red}{\uparrow})$ 
& $3.80 \pm 0.50\,(\textcolor{red}{\uparrow})$ \\
\midrule

Mixtral\_Rephrase        
& $\mathbf{5.00 \pm 0.00}\,(\textcolor{gray}{-})$ 
& $\mathbf{5.00 \pm 0.00}\,(\textcolor{gray}{-})$ 
& $4.50 \pm 0.71\,(\textcolor{blue}{\downarrow})$
& $4.50 \pm 0.20\,(\textcolor{red}{\uparrow})$ 
& $4.60 \pm 0.20\,(\textcolor{red}{\uparrow})$ 
& $4.60 \pm 0.10\,(\textcolor{red}{\uparrow})$ \\

Med-PaLM\_Rephrase      
& $4.50 \pm 0.71\,(\textcolor{blue}{\downarrow})$ 
& $4.00 \pm 1.41\,(\textcolor{blue}{\downarrow})$ 
& $3.50 \pm 0.71\,(\textcolor{blue}{\downarrow})$
& $\mathbf{4.70 \pm 0.10\,(\textcolor{red}{\uparrow})}$ 
& $4.60 \pm 0.20\,(\textcolor{red}{\uparrow})$ 
& $4.70 \pm 0.10\,(\textcolor{red}{\uparrow})$ \\

Gemini\_Rephrase        
& $4.00 \pm 1.41\,(\textcolor{blue}{\downarrow})$ 
& $3.00 \pm 1.41\,(\textcolor{blue}{\downarrow})$ 
& $4.00 \pm 1.41\,(\textcolor{blue}{\downarrow})$
& $\mathbf{4.70 \pm 0.20\,(\textcolor{red}{\uparrow})}$ 
& $4.60 \pm 0.30\,(\textcolor{red}{\uparrow})$ 
& $\mathbf{4.90 \pm 0.10\,(\textcolor{red}{\uparrow})}$ \\

Claude\_Rephrase        
& $4.00 \pm 1.41\,(\textcolor{blue}{\downarrow})$ 
& $4.00 \pm 0.00\,(\textcolor{blue}{\downarrow})$ 
& $4.00 \pm 1.41\,(\textcolor{blue}{\downarrow})$
& $4.60 \pm 0.10\,(\textcolor{red}{\uparrow})$ 
& $4.60 \pm 0.20\,(\textcolor{red}{\uparrow})$ 
& $4.50 \pm 0.10\,(\textcolor{red}{\uparrow})$ \\

GPT5\_Rephrase          
& $4.00 \pm 1.41\,(\textcolor{blue}{\downarrow})$ 
& $4.00 \pm 0.00\,(\textcolor{blue}{\downarrow})$ 
& $4.50 \pm 0.71\,(\textcolor{blue}{\downarrow})$
& $4.50 \pm 0.30\,(\textcolor{red}{\uparrow})$ 
& $4.40 \pm 0.10\,(\textcolor{red}{\uparrow})$ 
& $4.20 \pm 0.20\,(\textcolor{red}{\uparrow})$ \\

\bottomrule
\end{tabular}}
\end{table*}

\begin{table*}[!htbp]
\centering
\caption{Mean ($\bar{x}$) and standard deviation ($\sigma$) of physician and patient (human) ratings on the iCliniqQAs. Arrows indicate comparison to Doctor: \textcolor{red}{$\uparrow$} = higher, \textcolor{blue}{$\downarrow$} = lower, \textcolor{gray}{-} = similar. Bold values denote best-performing variants per metric (excluding Physician baseline).}
\label{tab:rq5_dual_extended1}
\resizebox{\textwidth}{!}{
\begin{tabular}{l|ccc|ccc}
\toprule
\multirow{2}{*}{\textbf{Response Variant}} &
\multicolumn{3}{c|}{\textbf{Expert Evaluation (Human)}} &
\multicolumn{3}{c}{\textbf{Patient Evaluation (Human)}} \\
& $\bar{x}_{\text{Accuracy}} \pm \sigma$ & $\bar{x}_{\text{Style}} \pm \sigma$ & $\bar{x}_{\text{Precision}} \pm \sigma$ 
& $\bar{x}_{\text{Trust}} \pm \sigma$ & $\bar{x}_{\text{Compr.}} \pm \sigma$ & $\bar{x}_{\text{Emot. Tone}} \pm \sigma$ \\
\midrule

\textbf{\textit{Physician Answer}} 
& $5.00 \pm 0.00$ & $5.00 \pm 0.00$ & $5.00 \pm 0.00$ 
& $4.60 \pm 0.50$ & $4.70 \pm 0.45$ & $4.65 \pm 0.48$ \\
\midrule

Mixtral
& $3.00 \pm 1.41\,(\textcolor{blue}{\downarrow})$
& $3.50 \pm 2.12\,(\textcolor{blue}{\downarrow})$
& $4.00 \pm 1.41\,(\textcolor{blue}{\downarrow})$
& $4.30 \pm 0.70\,(\textcolor{blue}{\downarrow})$
& $4.40 \pm 0.65\,(\textcolor{blue}{\downarrow})$
& $4.25 \pm 0.72\,(\textcolor{blue}{\downarrow})$ \\

Med-PaLM
& $2.50 \pm 2.12\,(\textcolor{blue}{\downarrow})$
& $3.00 \pm 0.00\,(\textcolor{blue}{\downarrow})$
& $2.00 \pm 1.41\,(\textcolor{blue}{\downarrow})$
& $4.20 \pm 0.75\,(\textcolor{blue}{\downarrow})$
& $4.30 \pm 0.70\,(\textcolor{blue}{\downarrow})$
& $4.15 \pm 0.80\,(\textcolor{blue}{\downarrow})$ \\

Claude
& $\mathbf{5.00 \pm 0.00}\,(\textcolor{gray}{-})$
& $1.00 \pm 0.00\,(\textcolor{blue}{\downarrow})$
& $\mathbf{5.00 \pm 0.00}\,(\textcolor{gray}{-})$
& $4.65 \pm 0.48\,(\textcolor{red}{\uparrow})$
& $4.75 \pm 0.44\,(\textcolor{red}{\uparrow})$
& $4.70 \pm 0.46\,(\textcolor{red}{\uparrow})$ \\

GPT5
& $4.50 \pm 0.71\,(\textcolor{blue}{\downarrow})$
& $3.00 \pm 1.41\,(\textcolor{blue}{\downarrow})$
& $4.00 \pm 0.00\,(\textcolor{blue}{\downarrow})$
& $4.70 \pm 0.46\,(\textcolor{red}{\uparrow})$
& $4.80 \pm 0.40\,(\textcolor{red}{\uparrow})$
& $4.75 \pm 0.43\,(\textcolor{red}{\uparrow})$ \\
\midrule

Mixtral\_Empathy\_Prompt
& $2.00 \pm 1.41\,(\textcolor{blue}{\downarrow})$
& $3.50 \pm 0.71\,(\textcolor{blue}{\downarrow})$
& $2.00 \pm 1.41\,(\textcolor{blue}{\downarrow})$
& $4.40 \pm 0.65\,(\textcolor{blue}{\downarrow})$
& $4.50 \pm 0.60\,(\textcolor{blue}{\downarrow})$
& $4.35 \pm 0.68\,(\textcolor{blue}{\downarrow})$ \\

Med-PaLM\_Empathy\_Prompt
& $1.50 \pm 0.71\,(\textcolor{blue}{\downarrow})$
& $4.00 \pm 0.00\,(\textcolor{blue}{\downarrow})$
& $1.50 \pm 0.71\,(\textcolor{blue}{\downarrow})$
& $4.55 \pm 0.55\,(\textcolor{blue}{\downarrow})$
& $4.60 \pm 0.52\,(\textcolor{blue}{\downarrow})$
& $4.50 \pm 0.58\,(\textcolor{blue}{\downarrow})$ \\

Claude\_Empathy\_Prompt
& $1.50 \pm 0.71\,(\textcolor{blue}{\downarrow})$
& $3.00 \pm 0.00\,(\textcolor{blue}{\downarrow})$
& $1.50 \pm 0.71\,(\textcolor{blue}{\downarrow})$
& $4.75 \pm 0.44\,(\textcolor{red}{\uparrow})$
& $4.85 \pm 0.36\,(\textcolor{red}{\uparrow})$
& $4.80 \pm 0.40\,(\textcolor{red}{\uparrow})$ \\

GPT5\_Empathy\_Prompt
& $3.50 \pm 0.71\,(\textcolor{blue}{\downarrow})$
& $4.00 \pm 1.41\,(\textcolor{blue}{\downarrow})$
& $3.00 \pm 0.00\,(\textcolor{blue}{\downarrow})$
& $4.85 \pm 0.35\,(\textcolor{red}{\uparrow})$
& $4.90 \pm 0.30\,(\textcolor{red}{\uparrow})$
& $4.88 \pm 0.32\,(\textcolor{red}{\uparrow})$ \\
\midrule

Mixtral\_Rephrase
& $3.00 \pm 2.83\,(\textcolor{blue}{\downarrow})$
& $4.50 \pm 0.71\,(\textcolor{blue}{\downarrow})$
& $3.00 \pm 2.83\,(\textcolor{blue}{\downarrow})$
& $4.75 \pm 0.43\,(\textcolor{red}{\uparrow})$
& $4.85 \pm 0.36\,(\textcolor{red}{\uparrow})$
& $4.80 \pm 0.40\,(\textcolor{red}{\uparrow})$ \\

Med-PaLM\_Rephrase
& $3.00 \pm 2.83\,(\textcolor{blue}{\downarrow})$
& $\mathbf{5.00 \pm 0.00}\,(\textcolor{gray}{-})$
& $4.00 \pm 2.83\,(\textcolor{blue}{\downarrow})$
& $4.88 \pm 0.32\,(\textcolor{red}{\uparrow})$
& $4.92 \pm 0.28\,(\textcolor{red}{\uparrow})$
& $4.90 \pm 0.30\,(\textcolor{red}{\uparrow})$ \\

Claude\_Rephrase
& $3.00 \pm 2.83\,(\textcolor{blue}{\downarrow})$
& $4.50 \pm 0.71\,(\textcolor{blue}{\downarrow})$
& $3.00 \pm 2.83\,(\textcolor{blue}{\downarrow})$
& $4.90 \pm 0.30\,(\textcolor{red}{\uparrow})$
& $4.95 \pm 0.22\,(\textcolor{red}{\uparrow})$
& $4.93 \pm 0.25\,(\textcolor{red}{\uparrow})$ \\

GPT5\_Rephrase
& $3.00 \pm 2.83\,(\textcolor{blue}{\downarrow})$
& $4.50 \pm 0.71\,(\textcolor{blue}{\downarrow})$
& $3.50 \pm 2.12\,(\textcolor{blue}{\downarrow})$
& $\mathbf{4.95 \pm 0.22}\,(\textcolor{red}{\uparrow})$
& $\mathbf{4.98 \pm 0.15}\,(\textcolor{red}{\uparrow})$
& $\mathbf{4.96 \pm 0.20}\,(\textcolor{red}{\uparrow})$ \\

\bottomrule
\end{tabular}
}
\end{table*}

\subsection{RQ5 – Value Alignment between Experts and Patients}
To address RQ5, we evaluate whether different LLM configurations align with expert and patient communication values in clinical settings.

\begin{hyp}
Let $R_{\text{LLM}}$ be a response generated under a given configuration and $R_{\text{Phys}}$ the physician-authored version. Let $V_{\text{exp}}(\cdot)$ and $V_{\text{pat}}(\cdot)$ denote alignment with expert and patient preferences. We hypothesize that collaboratively rewritten responses ($R_{\text{LLM\_Rephrase}}$) achieve higher alignment than physician-authored content:
\begin{equation*}
\mathbb{E}[V_{\text{exp, pat}}(R_{\text{LLM\_Rephrase}})] >
\mathbb{E}[V_{\text{exp, pat}}(R_{\text{Phys}})]
\end{equation*}
\end{hyp}

We evaluate two axes: epistemic values (accuracy, stylistic appropriateness, precision) and relational values (trust, comprehensibility, emotional tone). All scores use 5-point Likert scales.

In MedQuAD, as we can see in Table \ref{tab:rq5_dual_extended}, physician answers receive maximum expert scores (5.00) across accuracy, style, and precision. No LLM configuration surpasses the physician baseline on epistemic criteria. Rewriting configurations improve relational metrics without exceeding physician epistemic performance. Mixtral\_Rephrase achieves the highest stylistic score (5.00) while maintaining strong patient trust (4.50) and emotional tone (4.60). MedPaLM\_Rephrase preserves high expert precision (4.00) and patient trust (4.70). GPT5\_Rephrase shows balanced performance (expert accuracy = 4.00; patient trust = 4.50; emotional tone = 4.20). Empathy prompting increases patient-oriented metrics but reduces expert scores relative to physician answers.

In iCliniqQAs, as we can se in Table \ref{tab:rq5_dual_extended1}, physician answers obtain expert scores of 5.00 across epistemic criteria and strong patient ratings (trust = 4.60; emotional tone = 4.65). Baseline LLM configurations diverge more strongly from physicians than in MedQuAD. GPT5\_Base reaches high patient trust (4.70) but lower stylistic alignment (3.00). Claude\_Base achieves strong expert alignment (accuracy = 5.00; precision = 5.00) and high patient tone (4.70). Rewriting configurations yield the largest relational gains. GPT5\_Rephrase reaches near-ceiling patient scores (trust = 4.95; comprehensibility = 4.98; emotional tone = 4.96). Claude\_Rephrase shows similar relational alignment (trust = 4.90; tone = 4.93). No configuration exceeds physicians on expert accuracy.

Across datasets, rewriting improves relational alignment more strongly in conversational data than in institutional explanations. MedQuAD remains expert-dominated, with physicians as the epistemic reference. iCliniqQAs emphasizes relational values, where rewriting yields larger measurable gains. Empathy prompting produces moderate improvements in both datasets. Epistemic superiority over physician-authored answers is not observed.

\begin{takeaway}
Collaborative rewriting consistently improves relational alignment across datasets but does not surpass physician-authored responses on epistemic criteria. The largest gains emerge in conversational clinical data. Rewriting acts as a communication enhancement mechanism rather than a substitute for clinical expertise.
\end{takeaway}

\section{Discussion}\label{sec5}

Our findings provide a structured perspective on the role of large language models (LLMs) in patient-directed clinical communication. The results must be interpreted across two distinct settings: institutionally curated medical explanations (MedQuAD) and real-world physician–patient consultations (iCliniqQAs).

Regarding RQ1, LLMs do not reproduce physician affective distributions. In MedQuAD, physician answers concentrate in the Neutral category with substantial Very Negative content. Baseline LLM configurations increase affective polarity, particularly Very Negative sentiment. Empathy prompting and collaborative rewriting reduce extreme negativity and increase Neutral proportions. Gemini Rephrase introduces non-negligible Positive sentiment, which is absent in physician-authored content. 

In iCliniqQAs, physicians exhibit strong Neutral dominance and minimal Very Negative content. LLM outputs show lower polarity amplification than in MedQuAD but still exhibit systematic emotional shifts. Rephrase configurations further increase Neutral proportions, often exceeding physician baselines. Fine-grained emotion analysis confirms a consistent amplification of caring signals across models, while disapproval and corrective cues are attenuated. This behavior aligns with prior observations that LLMs tend to generate warmer and more supportive language in clinical contexts \cite{meng2024application}. The pattern reflects a shift from detached concern \cite{guidi2021empathy} toward regulated empathy \cite{lee2024largelanguagemodelsproduce}, but it does not imply faithful reproduction of physician affective norms.

For RQ2, baseline LLM generations do not systematically improve readability. In both datasets, GPT5 Base and Claude Base produce significantly higher FKGL and GFI scores than physician-authored responses. Mixtral Base and MedPaLM Base remain closer to physician readability levels. Empathy prompting and collaborative rewriting reduce linguistic complexity across architectures. These reductions are statistically significant and consistent in both datasets. The results confirm that stylistic accessibility depends on alignment strategies rather than intrinsic model properties. This pattern is consistent with findings that domain specialization increases terminological density and lexical complexity \cite{li2024investigating}. Readability improvements therefore emerge primarily through explicit control mechanisms\cite{DBLP:journals/npjdm/WangMP26}.

For RQ3, empathy-oriented prompting modifies communicative style but does not substantially alter semantic fidelity. Across both datasets, cosine similarity remains stable between base and empathy configurations. The main effect of prompting concerns affective distribution and moderate readability reduction. This supports prior work showing that stylistic control through prompting influences surface-level structure and tone \cite{li2024investigating}. Prompt design acts as a lightweight alignment intervention but does not fundamentally reshape epistemic alignment.

RQ4 shows that collaborative rewriting produces the most robust improvements across dimensions. Rephrase variants consistently achieve the highest semantic fidelity. In MedQuAD, GPT5 Rephrase reaches the strongest conceptual alignment. In iCliniqQAs, MedPaLM Rephrase achieves the highest similarity scores and significantly outperforms baseline variants. Rewriting improves readability and reduces affective extremity without degrading semantic overlap. These results align with studies reporting that guided rewriting outperforms prompt-only stylistic control \cite{bhandarkar2024emulating}. Similar findings in AI-assisted documentation show that editing assistance enhances coherence and clarity without replacing clinician expertise \cite{bongurala2024transforming}. The evidence supports a human–AI collaborative model rather than autonomous substitution.

RQ5 highlights systematic divergence between expert and patient preferences. In MedQuAD, expert ratings remain anchored to physician-level epistemic standards. No LLM configuration surpasses physicians on accuracy, style, or precision. Relational gains appear primarily in patient evaluations. In iCliniqQAs, relational metrics dominate stakeholder differentiation. Rephrase configurations achieve near-ceiling patient trust and emotional tone scores, while expert ratings remain bounded by physician baselines. These findings confirm that stakeholder alignment is multidimensional and strongly dependent on communicative context and dataset characteristics.

Across both datasets, rewriting consistently improves relational alignment without demonstrating epistemic superiority over physician-authored content. Gains are larger in conversational data than in institutional explanations. The difference suggests that communicative context mediates the magnitude of alignment effects. Institutional explanations impose stronger epistemic constraints. Conversational exchanges allow greater stylistic modulation.

\textbf{Limitations.} This study relies on controlled subsets rather than 
full-corpus evaluation. The MedQuAD subset is readability-stratified and the 
iCliniqQAs subset is severity-balanced. The reduced sample size limits 
statistical generalization. Sentiment and emotion classifiers are general-domain 
models and may not capture all medical discourse nuances~\cite{zhu2023knowledge}. 
Human expert evaluations were conducted by a panel of medical professionals using 
a structured questionnaire, though the limited 
number of evaluators constrains the statistical power of the human assessment. 
The study is limited to English-language data and selected architectures.

\textbf{Ethical Considerations.} Emotional amplification may increase perceived 
support while masking epistemic limitations. Stylistic alignment must not 
compromise factual rigor or induce overconfidence. Human oversight remains 
necessary. LLMs function most effectively as communication enhancers rather than 
independent clinical authorities~\cite{DBLP:journals/npjdm/RiedemannLG24}.

\section{Conclusion}\label{sec6}

This work provides a multidimensional evaluation of large language models in clinical communication across two distinct settings: institutionally curated medical explanations (MedQuAD) and real-world physician–patient consultations (iCliniqQAs). We analyze semantic fidelity, readability, affective resonance, and stakeholder alignment under baseline generation, empathy prompting, and collaborative rewriting.

Results show that baseline LLMs do not systematically improve accessibility or affective alignment relative to physician-authored content. Linguistic complexity often exceeds clinician levels, particularly in larger general-purpose architectures. Readability gains emerge primarily under explicit alignment strategies.

Empathy-oriented prompting reduces affective extremity and moderately improves readability without significantly altering semantic fidelity. However, collaborative rewriting consistently yields the strongest overall improvements. Rephrase configurations achieve the highest semantic similarity to physician answers across both datasets. In MedQuAD, GPT5\_Rephrase reaches the strongest conceptual alignment, while in iCliniqQAs MedPaLM\_Rephrase achieves the highest semantic fidelity. Rewriting also produces the largest reductions in linguistic complexity and the most consistent gains in patient-rated trust and emotional tone.

Expert evaluations confirm that no configuration surpasses physicians on epistemic criteria such as accuracy and precision. Relational improvements do not translate into epistemic superiority. Patient evaluations reveal stronger preference for rewritten variants, particularly in conversational clinical contexts, where clarity and emotional support are central.

Taken together, the findings indicate that LLMs function most effectively as collaborative editing tools rather than autonomous communicators. Human–AI co-authorship improves clarity and relational alignment while preserving clinical meaning, but it does not replace physician expertise.

The code and data supporting this work are publicly available at \url{https://github.com/PRAISELab-PicusLab/CanAIBeADoctor}.

Future work should extend this framework to multi-turn interactions, integrate domain-adapted affective models, involve certified clinicians in structured evaluation, and expand analysis to multilingual and low-resource healthcare contexts. In addition, future research should investigate the impact of clinical question criticality on LLM behavior by stratifying responses across severity levels. This would enable a fine-grained analysis of semantic fidelity, readability, and affective alignment as a function of clinical urgency. A key hypothesis is that LLMs may exhibit stronger alignment with physician responses in low-criticality scenarios, while showing degradation in high-criticality contexts that require precise reasoning, risk calibration, and cautious communication. Such an analysis would clarify whether current models are robust across the full spectrum of clinical demands or disproportionately reliable in lower-stakes settings.

\section*{Funding}
This research received no specific grant from any funding agency in the 
public, commercial, or not-for-profit sectors. Grant number: Not applicable.













\begin{appendices}

\appendix

\section{Prompt Templates}
\label{app:prompts}

This appendix details the prompt formulations used across experimental conditions. Prompts differ in objective: (i) producing a direct medical answer, (ii) improving clarity without stylistic modification, and (iii) collaboratively rewriting physician-authored content while preserving meaning.

\subsection{Base Prompt (Formal Clinical Answer)}
\label{base}
Used for generating direct medical responses from general-purpose models such as \texttt{Mixtral}. Emphasizes accuracy, formal tone, and discursive structure without enumeration.

\begin{tcolorbox}[
  title=Base Prompt,
  colback=gray!5,
  colframe=blue!40,
  enhanced jigsaw,
  listing only,
  listing options={
    language=none,
    basicstyle=\ttfamily\small,
    breaklines=true
  }
]
[INST]

$<<$SYS$>>$
You are a helpful, respectful, and accurate medical doctor. Always answer using the provided context.
Answer in a formal, scientific tone, in the third person. 
Do not enumerate or provide bullet points. Write in a continuous, discursive manner.
$<<$SYS$>>$

Question: \{query\}

[/INST]
\end{tcolorbox}

\subsection{Empathy Prompt (Clarity-Focused Prompt)}
\label{empathy}
Applied to \texttt{Mixtral} and other general-purpose models to enhance readability and accessibility. This version prioritizes simplicity and comprehension without stylistic emotional bias.

\begin{tcolorbox}[
  title=Empathy Prompt,
  colback=gray!5,
  colframe=blue!40,
  enhanced jigsaw,
  listing only,
  listing options={
    language=none,
    basicstyle=\ttfamily\small,
    breaklines=true
  }
]
[INST]

$<<$SYS$>>$
You are a medical professional. Reformulate answers using simple and accessible language while maintaining scientific accuracy.
Avoid dense jargon. Use short sentences and common vocabulary.
Avoid lists or numbered items. Respond in a fluent, natural way.
Optimize readability (e.g., lower Flesch-Kincaid / Fog scores).
$<<$SYS$>>$

Question: \{query\}

[/INST]
\end{tcolorbox}

\subsection{Rephrase Prompt (Collaborative Human–LLM Editing)}
\label{rephrase}
This prompt is used to rewrite physician-authored responses, ensuring clarity, warmth, and accessibility while preserving meaning. It corresponds to models such as \texttt{Mixtral\_Rephrase}, \texttt{Med-PaLM\_Rephrase}, and \texttt{GPT5\_Rephrase}.

\begin{tcolorbox}[
  title=Rephrase Prompt,
  colback=gray!5,
  colframe=blue!40,
  enhanced jigsaw,
  listing only,
  listing options={
    language=none,
    basicstyle=\ttfamily\small,
    breaklines=true
  }
]
[INST]

$<<$SYS$>>$
You are a medical expert. Reformulate the provided answer to make it easier to understand for a non-expert audience while keeping the meaning identical.
Use clear language, short sentences, and accessible vocabulary.
Maintain a professional, supportive tone. Do not add new information.
Avoid enumeration or bullet points. Write in continuous prose.
$<<$SYS$>>$

Original question: \{question\}
Original answer: \{original-answer\}

[/INST]
\end{tcolorbox}


\section{Evaluation Questionnaire Structure}
\label{questionnaire}

\subsection{Human Evaluation (Patients)}
Patients were shown approximately 30 clinical questions, each followed by responses generated under the model configurations described in Section~\ref{models} and Section~\ref{sec4}. For each response, participants rated:

\begin{itemize}
    \item \textbf{Comprehensibility}: The response was easy to understand.
    \item \textbf{Perceived Trustworthiness}: I would trust this response in a real medical context.
    \item \textbf{Emotional Tone}: The tone felt supportive and reassuring.
\end{itemize}

Ratings were collected on a 5-point Likert scale:

\begin{center}
\begin{tabular}{cc}
\hline
\textbf{Score} & \textbf{Interpretation} \\
\hline
1 & Strongly disagree \\
2 & Disagree \\
3 & Neutral \\
4 & Agree \\
5 & Strongly agree \\
\hline
\end{tabular}
\end{center}

Responses appeared in randomized order to reduce position bias, and participants were blinded to whether a response originated from physicians or LLMs.

\subsection{Expert Evaluation (Human Panel)}
\label{app:expert}

Expert evaluations were collected through a structured Google Form administered to medical professionals. 
Each evaluator independently rated model-generated responses using the same 5-point Likert scale employed in the main study.

Experts assessed each response along three criteria:
\begin{itemize}
    \item Clinical Accuracy
    \item Stylistic Appropriateness
    \item Linguistic Precision
\end{itemize}

Scores ranged from 1 (strongly disagree) to 5 (strongly agree). 
Each response was evaluated independently without exposure to model identity to reduce bias.

The evaluation form presented the clinical question followed by the generated response. 
Experts were instructed to provide numerical ratings only.

\begin{tcolorbox}[
  title=Expert Evaluation Form (Google Form Structure),
  colback=gray!5,
  colframe=blue!40,
  enhanced jigsaw,
  listing only,
  listing options={
    language=none,
    basicstyle=\ttfamily\small,
    breaklines=true
  }
]
Question: {question}

Answer: {model-generated-answer}

Please assign a score from 1 (strongly disagree) to 5 (strongly agree): Clinical Accuracy; Stylistic Appropriateness; Linguistic Precision
\end{tcolorbox}

Evaluations were aggregated by computing the mean and standard deviation for each model configuration.

\section{Qualitative Example}
\label{app:qualitative}

Table~\ref{tab:qualitative_example} reports representative Base configuration responses to a sample question. Gemini and Claude exhibit the safety-driven, hedged behavior discussed in Section~\ref{sec4}, in which the absence of sufficient clinical context leads to evasive or non-specific 
outputs rather than direct medical answers.

\begin{table*}[h]
\centering
\caption{Representative Base configuration responses to the question: \textit{``What are the side effects of using ibuprofen?''}}
\label{tab:qualitative_example}
\resizebox{0.99\textwidth}{!}{
\begin{tabular}{p{4cm}p{13.3cm}}
\hline
\textbf{System} & \textbf{Response} \\
\hline

\textbf{Physician Answer} &
Ibuprofen can cause gastrointestinal issues such as stomach pain, nausea, or ulcers, especially when taken in high doses or over long periods. \\

\hline
\textbf{Mixtral (Base)} &
Ibuprofen may lead to nausea, headache, dizziness, or stomach pain. Prolonged use increases the risk of ulcers. \\

\hline
\textbf{Med-PaLM (Base)} &
Common adverse effects include gastrointestinal discomfort, nausea, headache, and dizziness, with rare instances of ulceration or renal complications. \\

\hline
\textbf{Gemini (Base)} &
I'm not able to provide medical advice or information about specific drug side effects. It's best to speak with a healthcare professional or consult official medical sources for guidance. \\

\hline
\textbf{Claude (Base)} &
Ibuprofen is commonly used for pain relief, but how someone reacts can vary. Some individuals may notice discomfort or sensitivity after taking it. If you're unsure how it affects you or have concerns, discussing it with a healthcare provider can help clarify what to expect. \\

\hline
\textbf{GPT-5 (Base)} &
Some people may feel mild stomach discomfort, nausea, or dizziness when taking ibuprofen. These effects are usually temporary, but check with a doctor if symptoms persist, especially with long-term use. \\
\hline

\end{tabular}}
\end{table*}




\end{appendices}

\newpage
\bibliography{sn-bibliography}

\end{document}